\documentclass[10pt,twocolumn,letterpaper]{article}

\usepackage[pagenumbers]{cvpr} % To force page numbers, e.g. for an arXiv version

\usepackage{lipsum}  %added by dexter
\usepackage{graphicx}
\usepackage{amsmath}
\usepackage{amssymb}
\usepackage{multirow}
\usepackage{enumitem}
\usepackage{balance}
\usepackage{adjustbox}
\usepackage{array,ragged2e}

\usepackage{booktabs}
\usepackage[linesnumbered,ruled]{algorithm2e}
\SetAlgorithmName{Algorithm}{Algorithm}{Algorithm}
\IncMargin{0.5em}
\SetCommentSty{textnormal}
\SetNlSty{}{}{:}
\SetAlgoNlRelativeSize{0}
\SetKwInput{KwGlobal}{Global}
\SetKwInput{KwPrecondition}{Precondition}
\SetKwProg{Proc}{Procedure}{:}{}
\SetKwProg{Func}{Function}{:}{}
\SetKw{And}{and}
\SetKw{Or}{or}
\SetKw{To}{to}
\SetKw{DownTo}{downto}
\SetKw{Break}{break}
\SetKw{Continue}{continue}
\SetKw{SuchThat}{\textit{s.t.}}
\SetKw{WithRespectTo}{\textit{wrt}}
\SetKw{Iff}{\textit{iff.}}
\SetKw{MaxOf}{\textit{max of}}
\SetKw{MinOf}{\textit{min of}}
\SetKwBlock{Match}{match}{}{}

\DeclareMathOperator*{\argmax}{arg\,max}

\makeatletter
\DeclareRobustCommand{\iscircle}{\mathord{\mathpalette\is@circle\relax}}
\newcommand\is@circle[2]{%
  \begingroup
  \sbox\z@{\raisebox{\depth}{$\m@th#1\bigcirc$}}%
  \sbox\tw@{$#1\square$}%
  \resizebox{!}{\ht\tw@}{\usebox{\z@}}%
  \endgroup
}
\makeatother

\makeatletter
\newcommand{\xRightarrow}[2][]{\ext@arrow 0359\Rightarrowfill@{#1}{#2}}
\makeatother

\usepackage[dvipsnames]{xcolor}

\definecolor{cvprblue}{rgb}{0.21,0.49,0.74}
\usepackage[pagebackref,breaklinks,colorlinks,citecolor=cvprblue]{hyperref}

\def\paperID{3438} % *** Enter the Paper ID here
\def\confName{CVPR}
\def\confYear{2024}

\title{FER-C: Benchmarking Out-of-Distribution \\Soft Calibration for Facial Expression Recognition}

\author{Dexter Neo, Tsuhan Chen\\
School of Computing\\
National University of Singapore\\
{\tt\small e0534450@u.nus.edu,tsuhan@nus.edu.sg}
}
\begin{document}
\maketitle
%Appear well-calibrated but aren't, general calibration techniques do help
%%%%%%%%% ABSTRACT
\begin{abstract}
We present a soft benchmark for calibrating facial expression recognition (FER). While prior works have focused on identifying affective states, we find that FER models are uncalibrated. This is particularly true when out-of-distribution (OOD) shifts further exacerbate the ambiguity of facial expressions. While most OOD benchmarks provide hard labels, we argue that the ground-truth labels for evaluating FER models should be soft in order to better reflect the ambiguity behind facial behaviours. \textbf{Our framework proposes soft labels that closely approximates the average information loss based on different types of OOD shifts}. Finally, we show the benefits of calibration on five state-of-the-art FER algorithms tested on our benchmark. Our dataset and code is made available at: \textit{\href{https://anonymous.4open.science/r/FER-C-4197}{https://anonymous.4open.science/r/FER-C-4197}}. %Together our benchmark aims to aid future work towards FER networks that are safe and robust. Add to camera ready
\end{abstract}
%%%%%%%%% BODY TEXT
\section{Introduction}
Well-calibrated facial expression recognition is an important step towards achieving machines that can better understand human behaviour. Despite the tremendous advances in FER, it is often unclear whether modern methods are reliable enough for real-world deployment. The problem with FER can largely be attributed to flawed ground truth labels, since the labelling process is often noisy and subjective. Combined with the fact that facial expressions are highly ambiguous, low-quality images collected in-the-wild can further degrade the quality of annotations \cite{lukov_22}.

This makes obtaining reliable FER models that are not only accurate but well-calibrated especially challenging. Calibration aims to align a model's correctness with its confidence; however, when the provided ground-truth does not indicate any subjectivity or disagreement amongst annotators, it is difficult for practitioners to recognize whether models are overconfidently misclassifying samples or classifying them correctly with low confidence. 

%Prof says Fig 1 is obvious, need to highlight contributions how ?
\begin{figure}[!tb]
\centering
\includegraphics[width=\columnwidth]{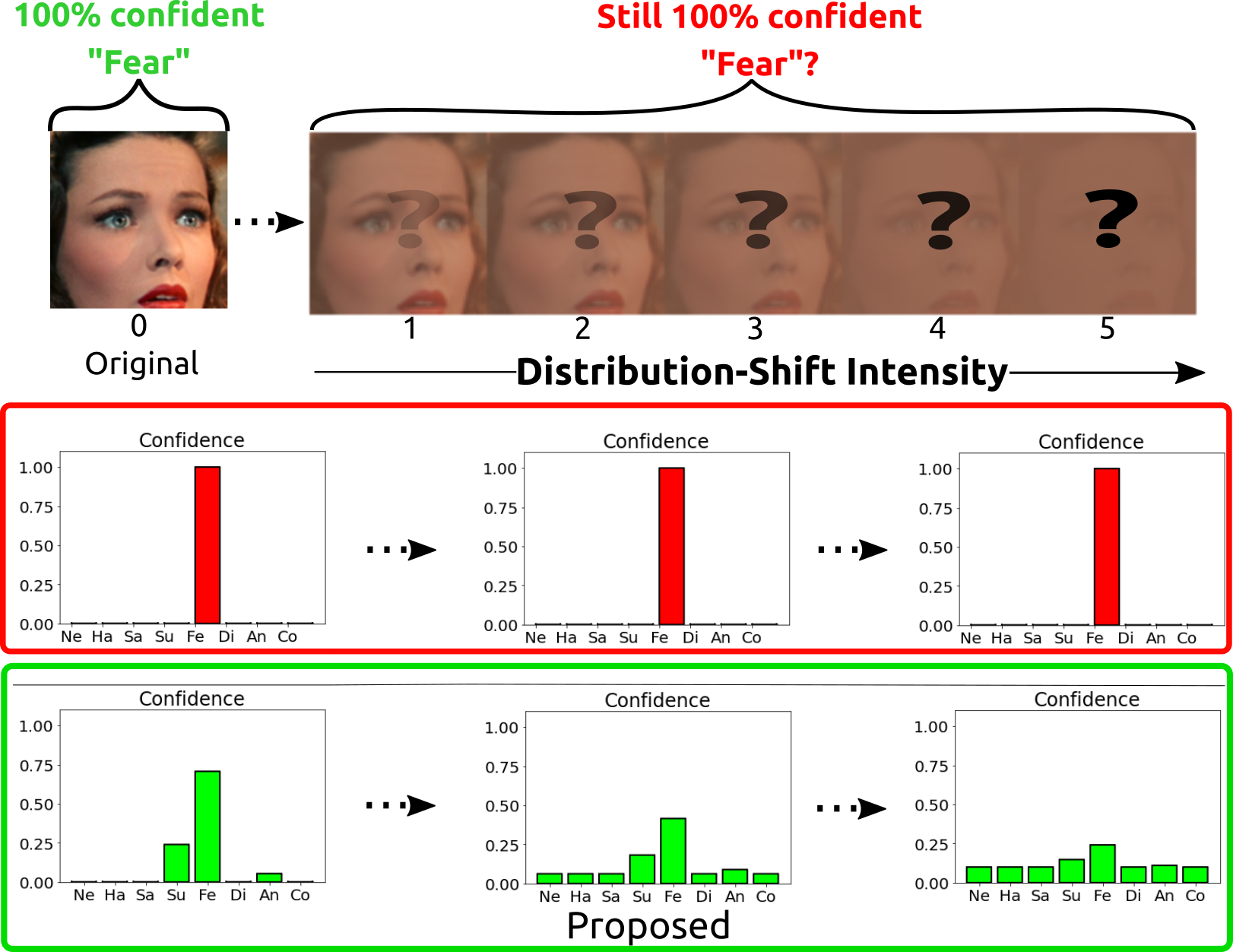}
\caption{As ambiguous facial expressions become increasingly OOD (top row), we propose softer/mixed test labels (lime) instead of one-hot labels (red). These test labels are softened based on the shift intensity and the likelihoods of next most probable classes.}
\label{fig:fig1}
\end{figure}
%Data points can be augmented, since end-users/annotators are human there is no perfect invariance guarantee. Dominant assumption is that augmented image $c(x_i)$ should retain 100\% of the given label. Can view proposed method as the average information loss resulting from the corruptions. Class specific smoothing -> highly occluded image of truck should raise the probabilities of other vehicles and not animal classes.

Untrustworthy probabilistic predictions from FER models can become dangerous when deployed in the real-world. Particularly in applications such as pain analysis \cite{ashraf_pain-07} or medical diagnosis \cite{Esteva2017DermatologistlevelCO, bandi2018detection}. Furthermore, FER models are not immune to out-of-distribution (OOD) shifts, such as potential changes in illumination, saturation, contrast, etc. These OOD shifts should lead to lower ground-truth confidence towards other classes and no longer exhibit high confidences (see \cref{fig:fig1}). 

The dominant assumption in OOD benchmarks \cite{hendrycks2019robustness, pmlr-v139-koh21a} is that augmented data points should preserve the original label with 100\% certainty. This assumption proves incompatible with fields such as FER where in-distribution (ID) expressions are rarely prototypically high-intensity and often appear compound/mixed \cite{Peng_2015_CVPR, zhou15, li2017reliable, li2019reliable}. This highlights the vital importance of a benchmark that not only covers the possible OOD shifts but also provides corresponding mixed ground-truth labels that reflects the subjectivity and increase in uncertainty. Ideally, we wish to have FER models that remain robust and well-calibrated despite inputs becoming OOD. Therefore, a soft calibration benchmark is a critical element towards the accurate assessment of the overall performance and uncertainty estimation of FER models. This leads to the question of how can we obtain meaningful soft labels that better reflect the uncertainties of facial expressions?

To this end, we propose a novel OOD FER benchmark derived from three of the most popular affect datasets. It consists of soft/mixed labels that correspond to standard image transformations similar to that of \cite{hendrycks2019robustness, deng2020labels}. We revisit concepts that \textit{non-uniformly} considers neighbouring classes and \textit{non-linearly} softens labels for increasingly shifted inputs. Our contributions can be summarized as follows:
%-------------------------------------------------------------------------
%Contributions
\begin{enumerate}
\item We present \textsc{FER-C}, corrupted OOD variants of three popular in-the-wild FER benchmarks: \textsc{AffectNet-C}, \textsc{Affwild-C} and \textsc{RAF-DB-C}.

\item Our benchmark mixes nearby classes based on their valence and arousal values, while also softening labels to account for OOD shifts.

\item We evaluate across five SOTA FER algorithms and show that DMUE and RUL performs best on both ID and OOD problem sets.

\item Finally, we show that calibration greatly benefits FER models - even when trained under label noise.
\end{enumerate}

\section{Related Work}
\subsection{Facial Expression Recognition}
Modern FER systems typically utilize deep learning, where either a convolutional neural network (CNN) or vision transformer (ViT) is trained end-to-end, acting as both a feature extractor and detector. This includes methods such as wide ensembles \cite{efficient}, de-expression learning \cite{Yang_2018_CVPR}, 3D morphable models \cite{Deep3DMM}, region attention networks \cite{wang2019region}, self-cure relabelling module \cite{wang2020suppressing}, factorized higher-order CNNs \cite{Kossaifi_2020_CVPR}, facial decomposition and reconstruction \cite{Ruan_2021_CVPR}, multi-task learning \cite{abaw_2023_CVPR}, learning from videos \cite{Zhang2023LearningER, Liu_2021_CVPR, Srivastava2023HowYF} and dynamic FER \cite{Lee_2023_CVPR, Wang_2023_CVPR}.

\subsection{Uncertainty in FER}
The ambiguity of facial expressions have led to the importance of alternative formulations apart from the basic Ekman model \cite{ekman1}. Since it is more common for emotions to appear compounded or mixed in day-to-day scenarios, many authors have proposed label distribution learning methods \cite{affective_2021, chen_CVPR_2020, she2021dive, zhang2021relative, Psaroudakis_2022_CVPR, lukov_22, Neo_2023_CVPR, Lee_2023_ICCV, wu2023net}. Others have proposed mixed FER databases with either multi-class or multi-label representations such as RAF-ML \cite{RAF-ML}, EMOTIC \cite{emotic_cvpr2017} and C-EXPR \cite{Kollias_2023_CVPR}. Current FER benchmarks lack the means to measure the performance of FER models when evaluated OOD. The main issue with obtaining mixed expression labels largely revolve around cost concerns, since manually annotating large datasets is expensive and time-consuming. Motivated by these issues, we propose an OOD, multi-class soft label benchmark without the use of manual labelling.

\begin{figure*}[ht]
\includegraphics[width=\textwidth]{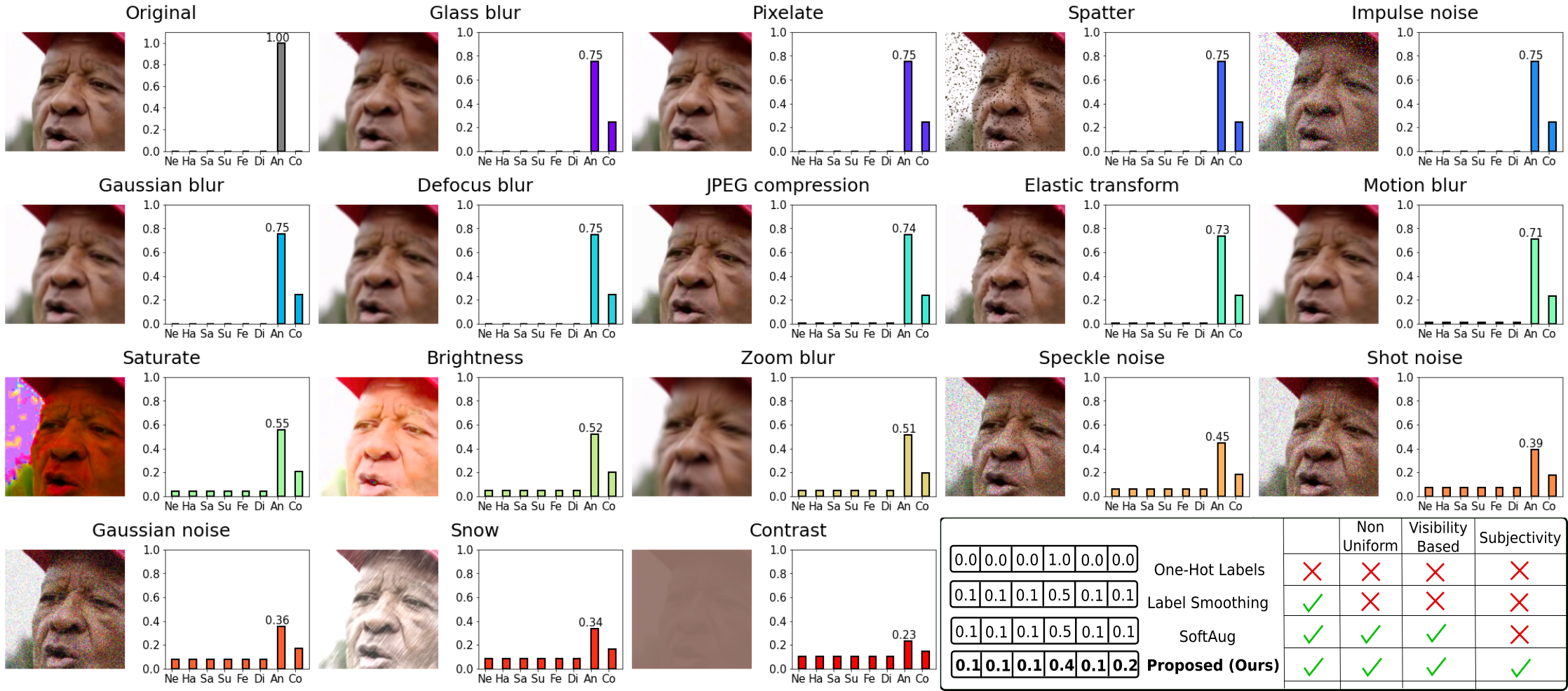}
\caption{Our proposed benchmark establishes soft labels corresponding to 17 different corruptions across five severity levels. For the AffectNet example shown here, the given label is \textit{'Anger'} with valence, arousal (-0.527, 0.663), which borders on both \textit{'Anger'} and \textit{'Contempt'}. Our test labels considers the likelihoods of neighbouring classes and are softened based on the visibility of each corruption.}
\label{fig:soft_corruption_labels}
\end{figure*}

\subsection{Calibration}
Modern neural network calibration techniques can be broadly summarized into two categories: 1.) Ad-hoc methods such as label smoothing \cite{Mller2019WhenDL}, temperature scaling \cite{Kull2019BeyondTS, yu2022robust, Joy2022SampledependentAT}, interpolation using splines \cite{gupta2021calibration}, soft label augmentation \cite{Liu_2023_CVPR}, adaptive class-wise smoothing \cite{liu2023cals}. 2.) Objective functions that include additional auxiliary terms with calibration and regularization properties such as Focal \cite{mukhoti2020calibrating}, InvFocal \cite{wang2021rethinking}, AvUC \cite{krishnan2020improving}, Soft-AvUC \cite{karandikar2021soft}, Poly\cite{leng2022polyloss} and MaxEnt loss \cite{neo2023maxent}. To measure the efficacy of these calibration methods various metrics have been proposed. We further discuss in detail the calibration problem in the following section.

\section{Preliminaries}
For atypical classification tasks, a model is trained with $N$ number of samples $(x_i, y_i)^N_{i=1}$ randomly drawn from a dataset $\mathcal{D}$. Where X, Y represents the input feature space, labels and $\mathcal{Y} = [1, 2, ..., K]$ is a fixed array of constants containing $K$ classes. For any input datum $x_i$ and training label $y_i$, the model is a deep neural network with learnable weights $\boldsymbol{\theta}$ and a softmax layer after the final fully connected layer with $K$ neurons. The model aims to predict the posterior distribution $\sum^K_{k=1} P_i(y|x) = 1$ from the given ground truth. The predicted top-1 class is obtained from the logits $\hat{y} := \argmax g^\theta_i(x)$, with the confidence $\hat{P}:= \max P_i(y|x)$. 

There exists a family of corruption functions $\mathcal{C}$ that represent the unknown real-world OOD shifts. In the case of OOD, the task is to predict the top class of the corrupted input $c(x_i)$ with confidence $\hat{P}:= \max P_i(y|c(x))$. Regardless of ID or OOD, a model is considered \textit{perfectly} calibrated iff the predicted probabilities are the same as the predicted likelihood of being correct, satisfying the definition for calibration $\mathbb{P}(\hat{y}=y| \hat{P}=P ) = P \quad \forall \in P[0,1]$. In practice, obtaining this definition of calibration is not possible unless the true posterior is known, therefore approximations are needed to estimate model calibration. In this paper, we measure model robustness and calibration on both ID and OOD problems sets using the following metrics.

\noindent \textbf{F1-Measure (F1):} The F1-measure or simply F1 is the harmonic mean between the Precision (PRC) and Recall or true positive ratio (TPR) of a classifier. For FER tasks, F1 is preferred over accuracy since it is common for imbalanced class distributions caused by certain expressions being rarer in-the-wild. The F1 score is given as:
\begin{equation}
\text{F1} = \frac{2 \times (\text{TPR} \times \text{PRC})}{\text{TPR} + \text{PRC}}.
\end{equation} 

\noindent \textbf{Negative Log-likelihood (NLL):} The NLL is also commonly known as the cross entropy in deep learning. Given a model's probabilities $P_i(y_k|x)$ and targets $y_{k}$, the NLL \cite{hastie01statisticallearning} is given as the mean of the summation across all $K$ classes: 
\begin{equation}
\text{NLL} = -\frac{1}{N}\sum^{N}_{i=1} \sum^{K}_{k=1} y_k \log{P_i(y_k|x)}.
\end{equation}

\noindent \textbf{Expected Calibration Error (ECE):} The ECE is a scalar that measures the difference between the expectations of confidence \textrm{conf} and accuracy \textrm{acc} \cite{10.5555/2888116.2888120}. By splitting the predictions into $B$ bins with each bin containing $n_b$ samples. The ECE is computed as the weighted absolute error between the \textrm{acc} and \textrm{conf} for each bin:
\begin{equation}
\text{ECE} = \sum^B_{b=1} \frac{n_b}{N} | \textrm{acc}(b) - \textrm{conf}(b)|.
\end{equation} The ECE is directly tied to the definition of calibration and converges to the B-term Riemann-Stieltjes sum \cite{pmlr-v70-guo17a, Tomani_2021_CVPR}.

\noindent \textbf{Kolmogorov-Smirnov Error (KSE):} 
As an alternative to the ECE, KSE numerically approximates the errors between two empirical cumulative distributions without any binning \cite{gupta2021calibration}. With the predicted probabilities $z_k$, the integral form of KSE is given by:
\begin{equation}
\text{KSE} = \int^1_0 |P(k|z_k) - z_k| P(z_k) dz_k
\end{equation}
We report results using accuracy, F1 for model robustness and NLL, ECE, KSE for model calibration. Similar conclusions are drawn when models are evaluated with other metrics such as AdaECE and CECE \cite{Nixon2019MeasuringCI}. For additional calibration metrics and results we refer readers to \cref{extra_metrics}.

\begin{figure*}[ht]
\includegraphics[width=\textwidth]{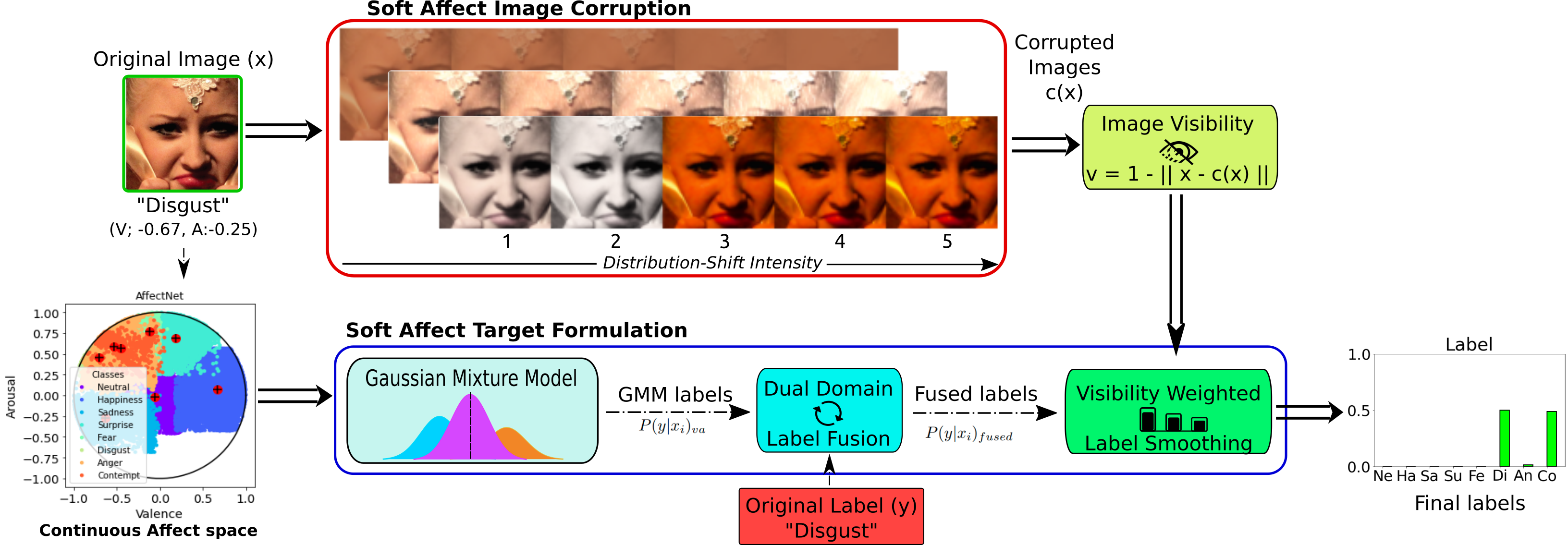}
\caption{Our framework formulates soft labels using Gaussian mixture models derived from the valence-arousal space. Combined with dual-domain label fusion and visibility weighted label smoothing we obtain soft likelihoods for each image under out-of-distribution shift.}
\label{fig:overall_framework}
\end{figure*}

\section{Soft Affect Benchmark}
\noindent
Summarized in \cref{fig:overall_framework}, our proposed test labels considers the likelihoods of neighbouring classes according to their valence-arousal (VA) values and are softened based on the visibility of each distribution shift type. \cref{soft_affect_image} discusses the benchmark design and image corruptions. \cref{soft_affect_target} justifies our proposed labels using adequately placed theoretical bounds such as Gaussian Mixture Models \cite{Neo_2023_CVPR} and human-visual studies \cite{Tang2017RecurrentCF}. 
\subsection{Soft Affect Image Corruption}
\label{soft_affect_image}
\subsubsection{Benchmark Design}
\textsc{FER-C} contains OOD versions of three popular in-the-wild FER datasets, namely \textsc{AffectNet-C}, \textsc{AffWild-C} and \textsc{RAF-DB-C}. A total of 17 different corruption types are applied to each dataset, with each corruption increasingly shifted OOD over five varying levels of severity/intensity (shown in \cref{fig:fig1} for the corruption \textit{Contrast}). These corruptions constitute a total of $17 \times 5 = 85$ variations for a given source image. In \cref{fig:soft_corruption_labels}, we show an example taken from the validation set of AffectNet augmented across the proposed 17 corruption types at severity level five. 

We apply these corruptions on the respective official validation/test sets of AffectNet, AffWild and RAF-DB. The corresponding corrupted images are saved on disk to ensure testing phases are fair and deterministic. As highlighted by \cite{hendrycks2019robustness}, these images are for testing only - \textit{not for training}. For comparisons between algorithms, all networks are trained using only clean images. During deployment, the best performing algorithm should be selected and retrained with data augmentations. In \cref{corruption_details}, we further discuss the details of each of the proposed corruptions. 

\subsection{Soft Affect Target Formulation}
\label{soft_affect_target}
\subsubsection{Gaussian Mixture Labels}
Given the nature of facial expressions, certain emotion classes are intrinsically closer to each other compared to others. For example, in \cref{fig:overall_framework} we show how the given \textit{'Disgust'} expression may also be interpreted as \textit{'Contempt'}. In order to capture this subjectivity of expressions, we incorporate a probabilistic approach in our soft label design by non-uniformly considering the likelihoods of neighbouring classes based on the VA annotations \cite{Neo_2023_CVPR}. Consider the following Gaussian Mixture Model (GMM) \cite{bishop:2006:PRML} from the continuous affect space \cite{circumplex}:
\begin{equation}
P(x_i^{\text{VA}}) = \sum^K_{k=1}\pi_k\mathcal{N}(x_i^{\text{VA}}|\mu_{k},\Sigma_{k})
\end{equation}
The GMM is a complex probability density function consisting of a linear superposition of $K$ Gaussians for each class. Where $x_i^{\text{VA}} \in \{-1, +1\}$ represents the valence, arousal values for the randomly drawn sample $x_i$. The marginal distribution/normalizer $\mathcal{N}(x_n|\mu_k, \Sigma_k)$ for each Gaussian is defined by the prior distribution $\pi_k$, centroids $\mu_k$ and covariance matrices $\Sigma_k$. The parameters describing each Gaussian component are computed within one standard deviation from the VA annotations in the training set. From the GMM, an estimate of the posterior distribution $P(y|x_i)_{\text{VA}}$ for each sample can be obtained using Bayes' Theorem: 
\begin{equation}
P(y|x_i)_{\text{VA}} = \frac{\pi_k\mathcal{N}(x_i^{\text{VA}}|\mu_{k},\Sigma_{k})}
{\sum^K_{j=1}\pi_j\mathcal{N}(x_i^{\text{VA}}|\mu_{k},\Sigma_{k})}
\label{GMM_labels}
\end{equation}
where each VA value maps to a set of valid probabilities $\sum_k P(y|x_i)_{\text{VA}} = 1$. Transitioning from one Gaussian component to another would result in different posterior estimates.
\subsubsection{Dual-Domain Label Fusion}
After obtaining GMM probabilities $P(y|x_i)_{\text{VA}}$, naively using these soft targets could result in a loss of training accuracy. Since the VA annotations are not immune to the noise and subjectivity of facial expressions, certain images may have been wrongly annotated \cite{Neo_2023_CVPR}. To circumvent this issue, we propose to fuse the given ground-truth one-hot labels $y_k$ with the derived distributions $P(y|x_i)_{\text{VA}}$ as follows:
\begin{equation}
P(y|x_i)_{\text{fused}} = \beta y_k + (1 - \beta) P(y|x_i)_{\text{VA}}
\label{fusion_equation}
\end{equation}
$\beta$ is used as a hyperparameter controlling the degree of contribution between the discrete and continuous labels. Recent studies \cite{wang2018twolevel, kollias2020face, Le_2023_WACV, Neo_2023_CVPR} support the relationships between discrete and continuous affect spaces. By performing label fusion, we encourage better synergy between both models of affect.

\subsubsection{Visibility Weighted Label Smoothing}
\begin{figure}[!t]
\centering
\includegraphics[width=\columnwidth]{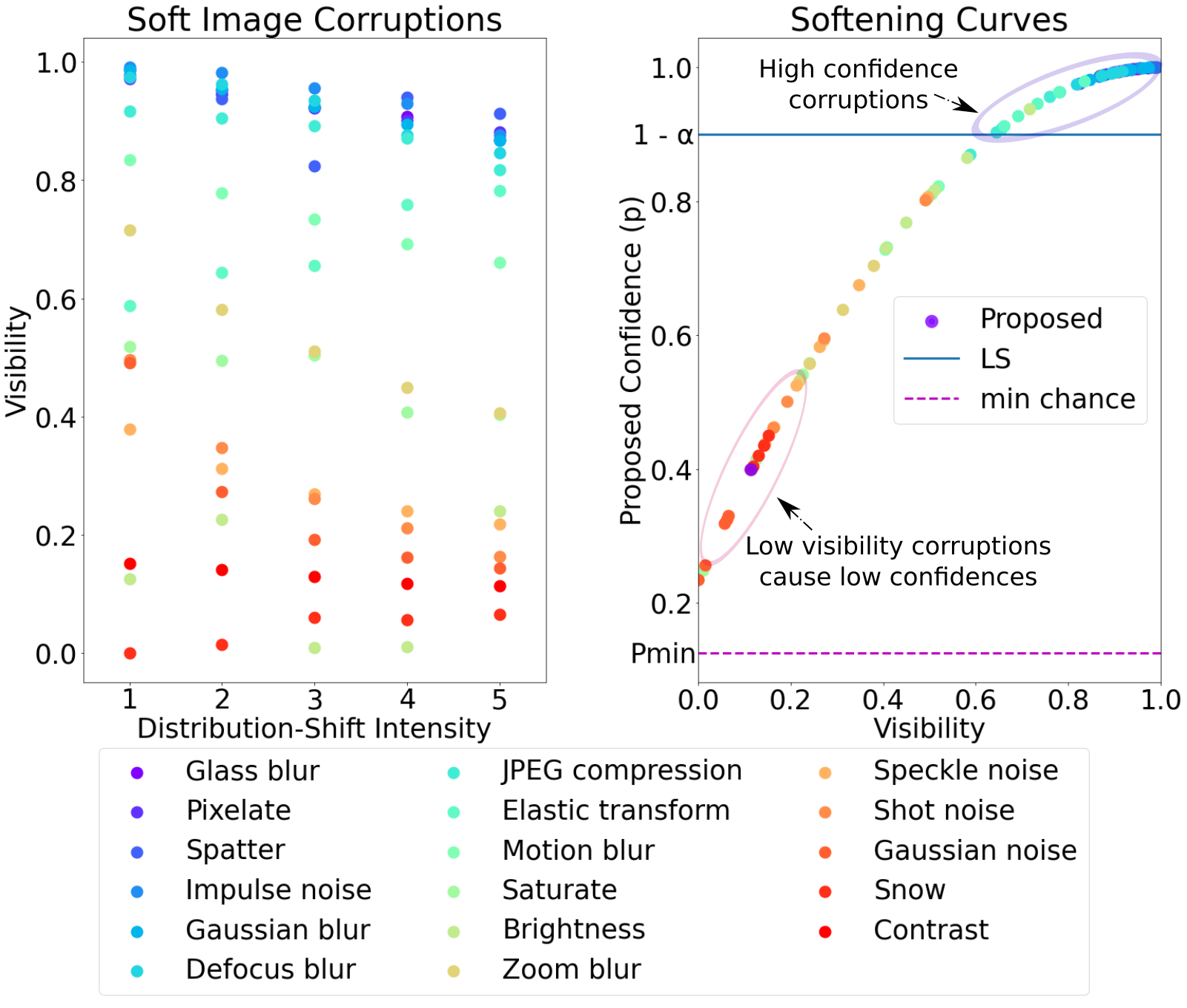}
\vspace{-1.5em}
\caption{Left: Normalized visibilities of the 85 corruptions. Right: Non-linear mapping of each corruption/visibility to softened target confidences. Mild corruptions lead to higher confidences, while severe corruptions cause lower confidences.}
\label{fig:ood_error}     
\end{figure}

To complete the derivation of soft targets for each OOD image $x_i$, we propose an intuitive approach to soften the labels based on the visibility caused by the corruption. Mathematically, we aim to find a suitable label for the corrupted image $c(x_i)$:
\begin{equation}
(x_i, y_i) \xRightarrow[\text{distribution shift}]{} (c(x_i), s_i) 
\end{equation}
Prior works \cite{Tang2017RecurrentCF, Liu_2023_CVPR} suggest that certain cases of mild occlusions are perceptually invariant to the human eye. In other extreme cases, corruptions can cause 100\% information loss. Consider the following definition of a general visibility function, which measures the average degradation of image quality:
\begin{equation}
    v_i = 1 - ||x_i - c(x_i)||_2
\label{visibility}
\end{equation}
where $v_i \in [0-1]$ is the mean normalized L2 difference of the current corruption divided by the worst corruption. On the left side of \cref{fig:ood_error}, we show the scatterplot of the average visibility for 100 random samples and each of their 85 corrupted forms. Lower visibility corresponds to poorer image quality, resulting in higher uncertainty and lower confidences. Notably, the greatest drops in visibility are caused by corruptions such as \textit{Snow} and \textit{Contrast}. Conversely, \textit{Impulse noise} or \textit{Glass blur} does not cause much visual information loss.

%Addressing the criticsm of using MSE for image quality assesment, SSIM is sensitive to luminance, contrast and structural changes. In our experiments, we found MSE to be more reliable than SSIM

%In most cases, increasing the distribution shift intensity would lead to lower visibilities. However due to the nature of images collected in-the-wild, effects of certain corruptions can differ, improving visibility with greater OOD shift. Consider an image with a darker backdrop, increasing its \textit{'Contrast'} would actually improve the image content.

\begin{figure*}[!htb]
\small
\centering
\includegraphics[width=\textwidth]{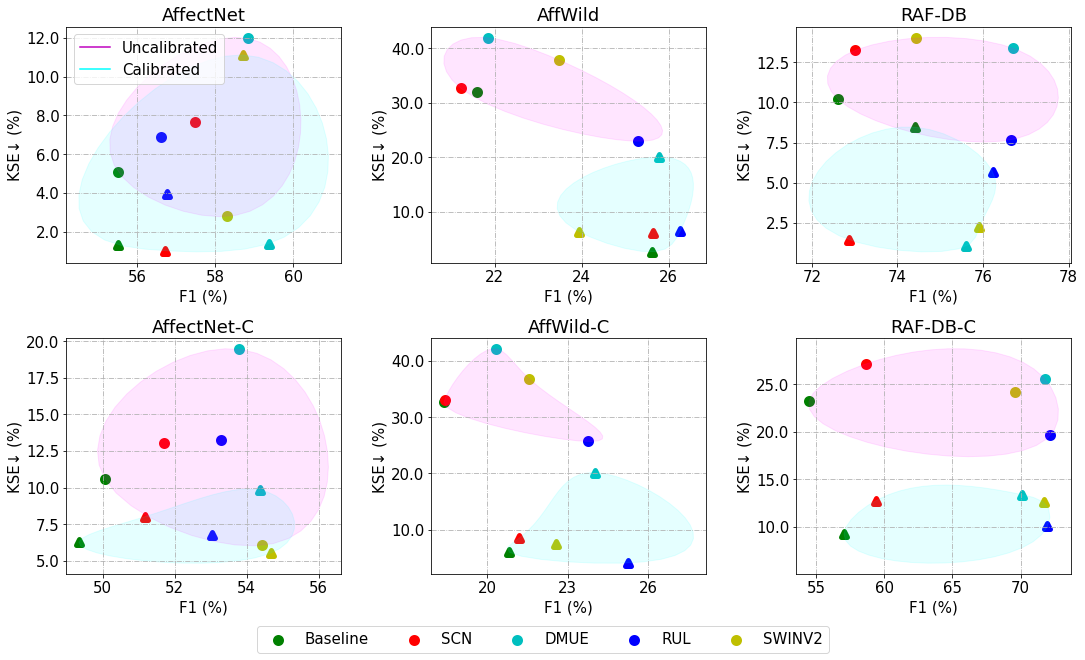}
\vspace{-6mm}
\caption{\textbf{Top}: Calibration improves the ECE scores of FER methods for most ID cases. \textbf{Bottom}: When evaluated OOD, calibration greatly benefits all FER methods. $\iscircle$ and $\triangle$ respectively denotes the calibrated and uncalibrated forms of each method}.
\label{fig:scatter_plots}
\end{figure*}
Following SoftAug \cite{Liu_2023_CVPR}, we propose to use a non-linear label smoothing approach based on the normalized visibilities of each corruption. Label smoothing \cite{Mller2019WhenDL} softens the target distribution by \textit{uniformly} redistributing the probabilities of the given label to other classes. Specifically, the formula for label smoothing is given by: $s_i = (1 - \alpha) y_k + \frac{\alpha}{K}$, where $\alpha$ is a handcrafted hyperparameter and $s_i$ is the resultant smoothed label. We reformulate the smoothing parameter $\alpha$ such that it is dependant on the image visibility $v_i$ and an exponent $\kappa$.

\begin{equation}
    \alpha(v_i, \kappa) = (1 - P_{min}) (1 - v_i)^\kappa
\label{soft_aug}
\end{equation}
where $P_{min}$ is the minimum probability or uniform prior, \eg $P_{min} = \frac{1}{K} = 0.125$ for AffectNet. On the right side of \cref{fig:ood_error}, we show the non-linear softening curves which maps each corruption to a target confidence. Corruptions such as \textit{Glass blur} have high visibility and retain higher confidences, whereas corruptions like \textit{Contrast} have lower visibility and lead to the lower confidences (see \cref{fig:soft_corruption_labels}). A non-linear softening of targets better captures the degradation of human vision caused by occlusions \cite{Tang2017RecurrentCF, Liu_2023_CVPR}, whereas label smoothing only applies a fixed softening factor. This visual degradation is best captured when $\kappa=2$. Combined with the aforementioned GMM and label fusion modules, the final label is given as:
\begin{equation}
    P(y|x_i)_{\text{final}} = (1 - \alpha(v_i, \kappa)) P(y|x_i)_{fused} + \frac{\alpha(v_i, \kappa)}{K}
\label{final_label}
\end{equation}

We note that alternative formulations of the visibility function are possible, such as SSIM \cite{SSIM} which measures the structural degradadation rather than absolute pixelwise difference.  However, we found the L2 difference to be a better measure of image quality than SSIM for our soft label formulation (see \cref{image_metrics}). Additionally, we also conduct a user-study to verify the proposed soft labels as a sanity check, with more details discussed in \cref{user_study}.

\section{Experiments and Results}
We conduct inhouse experiments on the following three FER datasets and apply our proposed framework to generate their OOD variants. We evaluate the performances of the following five SOTA FER algorithms: a baseline \cite{affectnet}, SCN \cite{wang2020suppressing}, DMUE\cite{she2021dive}, RUL \cite{zhang2021relative} and SwinV2-T\cite{liu2021swinv2}. We further discuss the performances of each algorithm before and after calibration on ID, OOD and label noise experiments. For additional details regarding experiments and hyperparameter settings, we refer readers to \cref{implementation_details}. %Our benchmark can be viewed anonymously at \textit{\href{https://github.com/dexterdley}{https://github.com/dexterdley/RC-AffectNet} }.

\subsection{Datasets}
\textbf{AffectNet} \cite{affectnet} consists of 450K manually annotated images with both discrete and continuous labels. As the test set is unreleased, we randomly split a subset of training images for validation and hyperparameter selection. The official validation set of 4K images is used for testing and augmented into \textsc{AffectNet-C}.

\noindent \textbf{AffWild} \cite{zafeiriou2017aff, kollias2019expression} is the largest FER video database. To generate GMM labels for each video frame, the parameters $\pi_k, \mu_k, \Sigma_k$ are borrowed from AffectNet. We create a makeshift image dataset by performing an 80/20 train/test random split based on the unique identities from the 252 AffWild training videos. Which gives 240 images per identity or 54K training and 4.8K test images with both VA and discrete expression annotations. %We are unable to use AffWild2 \cite{kollias2019expression} since the VA and class labels do not overlap with the videos (i.e. both label types are provided, but not for the same videos.) 

\noindent \textbf{RAF-DB} \cite{li2017reliable, li2019reliable} includes 30K facial images manually annotated single and multi-class labels. As RAF-DB already contains mixed labels, we did not use the GMM formulation in \cref{GMM_labels}. Instead, we directly apply image corruptions and visibility weighted label smoothing to create \textsc{RAF-DB-C}.

\begin{table*}[!htb]%\setlength\tabcolsep{0.25em}
\centering
\large
\begin{adjustbox}{width=\textwidth}
\begin{tabular}{l|cccc|cccc|cccc}
&\multicolumn{4}{c|}{(a) AffectNet} &\multicolumn{4}{c|}{(b) AffWild} &\multicolumn{4}{c}{(c) RAF-DB}\\
Method &F1 (Pre) &ECE (Pre) &F1 (Post) &ECE (Post) &F1 (Pre) &ECE (Pre) &F1 (Post) &ECE (Post) &F1 (Pre) &ECE (Pre) &F1 (Post) &ECE (Post)\\
\hline
Baseline\cite{affectnet} &56.4 $\pm{0.9}$ &6.10$\pm{2.7}$ &55.8$\pm{0.3}$ &\underline{4.00$\pm{1.0}$} &20.8$\pm{0.8}$ &33.7$\pm{1.7}$ &25.1$\pm{0.5}$ &\underline{\textbf{7.37$\pm{0.8}$}} &72.6$\pm{0.1}$ &9.86$\pm{0.4}$ &73.8$\pm{0.6}$ &\underline{7.51$\pm{1.0}$} \\
SCN\cite{wang2020suppressing} &56.9 $\pm{0.5}$ &6.75$\pm{3.1}$ &56.7$\pm{0.3}$ &\underline{4.71$\pm{2.9}$} &20.5$\pm{0.7}$ &33.5$\pm{0.9}$ &24.9$\pm{0.8}$ &\underline{10.1$\pm{0.3}$} &73.4$\pm{0.4}$ &11.7$\pm{1.6}$ &73.0$\pm{0.1}$ &\underline{4.73$\pm{0.9}$} \\
DMUE\cite{she2021dive} &58.8 $\pm{0.0}$ &14.4$\pm{2.3}$ &\underline{\textbf{58.7$\pm{0.7}$}} &\underline{\textbf{4.67$\pm{2.5}$}} &24.6$\pm{2.7}$ &41.1$\pm{0.8}$ &25.5$\pm{0.2}$ &\underline{18.9$\pm{1.1}$} &\underline{\textbf{76.5$\pm{0.2}$}} &13.3$\pm{0.1}$ &75.8$\pm{0.2}$ &\underline{5.60$\pm{1.8}$} \\
RUL\cite{zhang2021relative} &57.3 $\pm{0.6}$ &7.20$\pm{0.4}$ &57.3$\pm{0.5}$ &\underline{5.63$\pm{0.3}$} &25.2$\pm{0.1}$ &22.6$\pm{0.4}$ &\underline{\textbf{26.7$\pm{0.4}$}} &\underline{8.89$\pm{0.4}$} &75.6$\pm{1.0}$ &6.15$\pm{1.9}$ &76.1$\pm{0.1}$ &\underline{4.78$\pm{0.8}$} \\
SWINV2\cite{liu2021swinv2}&57.8 $\pm{0.5}$ &\underline{3.63$\pm{0.7}$} &57.9$\pm{0.8}$ &9.89$\pm{1.3}$ &21.9$\pm{1.6}$ &39.0$\pm{1.1}$ &22.5$\pm{1.4}$ &\underline{11.8$\pm{1.5}$} &74.6$\pm{0.2}$ &13.8$\pm{0.2}$ &75.2$\pm{0.7}$ &\underline{\textbf{4.36$\pm{0.2}$}} \\
&\multicolumn{4}{c}{(d) AffectNet-C} &\multicolumn{4}{c}{(e) AffWild-C} &\multicolumn{4}{c}{(f) RAF-DB-C}\\
Method &F1 (Pre) &ECE (Pre) &F1 (Post) &ECE (Post) &F1 (Pre) &ECE (Pre) &F1 (Post) &ECE (Post) &F1 (Pre) &ECE (Pre) &F1 (Post) &ECE (Post)\\
\hline
Baseline\cite{affectnet} &50.4$\pm{0.3}$ &12.3$\pm{0.9}$ &49.7$\pm{0.3}$ &\underline{8.36$\pm{0.5}$} &18.0$\pm{0.3}$ &32.7$\pm{0.4}$ &20.7$\pm{1.4}$ &\underline{9.68$\pm{1.0}$} &55.2$\pm{0.7}$ &23.2$\pm{0.4}$ &58.2$\pm{1.1}$ &\underline{\textbf{11.1$\pm{0.1}$}} \\
SCN\cite{wang2020suppressing} &51.2$\pm{0.4}$ &12.6$\pm{2.8}$ &50.9$\pm{0.3}$ &\underline{9.22$\pm{1.5}$} &18.1$\pm{0.3}$ &32.9$\pm{0.6}$ &20.7$\pm{0.4}$ &\underline{11.7$\pm{1.1}$} &58.3$\pm{0.4}$ &25.0$\pm{2.4}$ &59.4$\pm{0.1}$ &\underline{13.3$\pm{1.4}$} \\
DMUE\cite{she2021dive} &54.5$\pm{0.8}$ &22.0$\pm{1.5}$ &\underline{\textbf{54.6$\pm{0.3}$}} &\underline{13.3$\pm{1.9}$} &22.6$\pm{2.3}$ &42.0$\pm{0.3}$ &23.5$\pm{0.5}$ &\underline{20.1$\pm{0.5}$} &71.7$\pm{0.1}$ &25.5$\pm{0.1}$ &70.6$\pm{0.5}$ &\underline{17.3$\pm{2.7}$} \\
RUL\cite{zhang2021relative} &53.7$\pm{0.4}$ &15.7$\pm{0.2}$ &53.6$\pm{0.6}$ &\underline{10.5$\pm{0.2}$} &23.3$\pm{0.5}$ &26.3$\pm{0.6}$ &\underline{\textbf{25.1$\pm{0.1}$}} &\underline{\textbf{8.03$\pm{0.2}$}}  &71.4$\pm{0.8}$ &18.2$\pm{2.1}$ &\underline{\textbf{72.0$\pm{0.1}$}} &\underline{13.2$\pm{0.4}$} \\
SWINV2\cite{liu2021swinv2} &54.0$\pm{0.4}$ &11.3$\pm{1.7}$ &54.0$\pm{0.6}$ &\underline{\textbf{7.54$\pm{0.1}$}} &20.5$\pm{1.0}$ &37.6$\pm{0.6}$ &21.3$\pm{1.2}$ &\textbf{11.5$\pm{1.6}$} &70.2$\pm{0.6}$ &24.2$\pm{0.1}$ &71.0$\pm{0.7}$ &\underline{14.6$\pm{0.6}$} \\
\end{tabular}
\end{adjustbox}
\vspace{-3mm}
\caption{
Evaluation results (\%) for ID (top) and OOD (bottom) test sets averaged over 3 seeds. Calibration greatly benefits FER models with no significant change in recognition performance or training time. All methods are trained with \textit{clean} images and one-hot labels.}
\label{table:main}
\end{table*}

\subsection{Method and Calibration Strategy}
We show the benefits of calibration by retraining all algorithms using
a combination of methods from the calibration literature: MaxEnt loss \cite{neo2023maxent}, which is a extension of Focal loss \cite{focal,mukhoti2020calibrating} designed for handling OOD inputs and Margin-based Label Smoothing (MBLS) \cite{liu2022mbls}, which restricts the maximum output logit within a user-defined margin. Both algorithms are state-of-the-art calibration strategies and can be used in tandem with only minor code modifications. Specifically, the MaxEnt loss is given by:
\begin{equation}
\begin{split}
&\mathcal{L}_{\text{ME}} = \underbrace{-\sum_k \bigl(1 - P_i(y_k|x_i) \bigr)^\gamma \log P_i(y_k|x_i) }_{\text{Focal Loss}} \\ 
& \qquad + \lambda_\mu \Biggl[ \underbrace{\sum_k \mathcal{Y} P_i(y_k|x) - \mu_G}_{\text{Global mean constraint}}  \Biggr]
\label{expo_final}
\end{split}
\end{equation}
which extends the Focal loss by constraining the predicted posteriors to the observed expectation. For \eg assuming a uniform prior of 8 classes, the global expected value is computed as $\sum^{K=8}_{k=1} \mathcal{Y}P(\mathcal{Y})=\mu_G=3.6$. $\lambda_\mu$ controls the strength of the constraint and $\gamma \ge 0$ controls the Shannon entropy term. Next, the MBLS is given by:
\begin{equation}
\mathcal{L}_{\text{MBLS}} = \sum_k \max(0, \max_j(g_j^\theta(x_i)) - g_k^\theta(x_i) - m)
\end{equation}
Where the margin is recommend by the authors to be fixed at $m=10$ for image tasks \cite{liu2022mbls}. In our experiments we have tried CALS-ALM \cite{liu2023cals}, but have found CALS-ALM to only improve calibration ID and MBLS to be more reliable OOD. Therefore, the final objective function is given by:
\begin{equation}
\mathcal{L} = \mathcal{L}_{\text{ME}} + \mathcal{L}_{\text{MBLS}}
\label{final_objective_func}
\end{equation}

\subsubsection{In-Distribution Performance}
For ID experiments, each algorithm is tested on clean images and the original one-hot labels. In \cref{fig:scatter_plots}, we cluster both uncalibrated (magenta) and calibrated (cyan) forms of each algorithm's test KSE and F1. Our findings suggests that most methods remain relatively well-calibrated on AffectNet and calibration provides a moderate improvement to all algorithms except SWINV2. Compared to past convolutional architectures \cite{minderer2021revisiting}, vision transformers tend to be better calibrated and more robust, explaining SWINV2's low ECE scores on AffectNet without calibration. However, on RAF-DB and AffWild calibration clearly benefits all algorithms with their calibrated forms having lower KSE. 

We further support our findings in Table \ref{table:main} by showing the average F1 and ECE scores across 3 random seeds. We compute ECE using 15 bins to mitigate any form of bias towards larger bins and show the best calibration scores underlined with $\pm$ indicating 1 standard deviation. We can see that the most consistent algorithm performance comes from DMUE with typically high F1 (59\%, 26\%, 77\%) paired by low ECE (4.6\%, 19\%, 5.6\%) respectively on all 3 datasets. We highlight that for the ID performance, the scores reported in our paper are lower than those reported by the respective authors. This is because we train all algorithms without any image transformations. With additional on-the-fly augmentations used during training, we match the original reported scores.

\subsubsection{Out-of-Distribution Performance}
The benefits of training well-calibrated FER models become more obvious when evaluated OOD. For OOD performance, we report the test scores on \textsc{FER-C} (average of all five distribution shifts) and proposed soft labels. In \cref{fig:scatter_plots}, all methods continue to exhibit lower KSE scores post-calibration as compared to methods without calibration. We also find the effects of calibration more pronounced on the OOD set, with larger improvements in KSE and ECE. Compared to their ID test sets, we roughly observe a (5\%, 3\% and 10\%) drop in F1 caused by the OOD shifts for each dataset. Coincidentally, DMUE is again able to obtain the best performance overall, with F1 (54\%, 24\%, 71\%) and ECE (82\%, 16\%, 15\%), with RUL and SWINV2 coming in second and third. 

Ideally, we wish to obtain a clear margin of separation between calibrated and uncalibrated clusters for both ID and OOD sets. However, this requires a rigourous selection of hyperparameters such as $\gamma$ and $m$ on a separate validation set. Which remains a key issue for model calibration as the validation set may not be ID with the test set \cite{Ovadia2019CanYT}. This also means that hyperparameters selected on one dataset may not translate well to other datasets or deployment. In general, we urge practitioners to use calibration algorithms/loss functions that can generalize well to different settings. %Some good choices include our proposed strategy or Focal loss which delivers accurate and well-calibrated models with no added training times.
\begin{figure}[!b]
\centering
\includegraphics[width=\columnwidth]{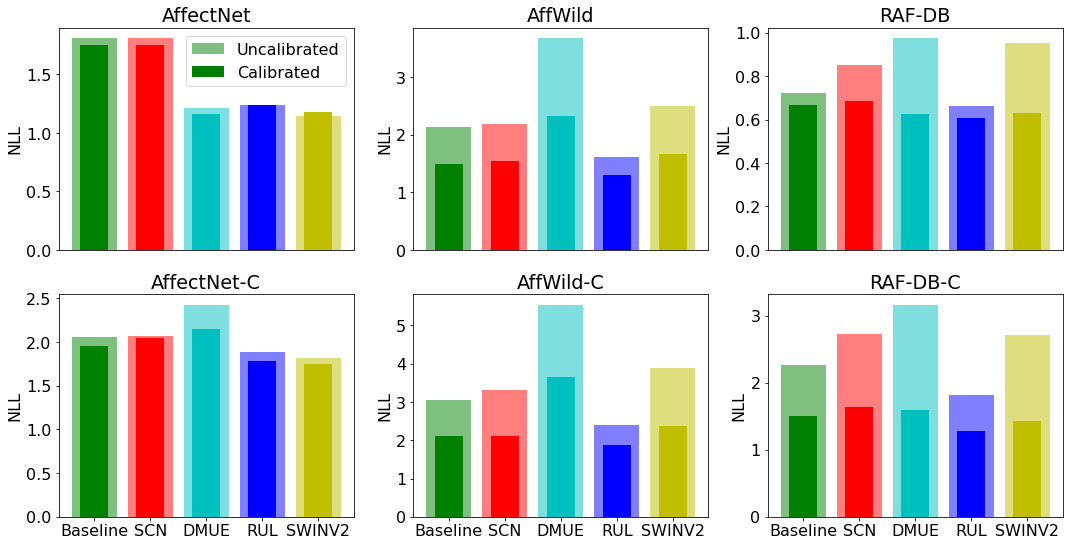}
\vspace{-2em}
\caption{Calibrated FER models demonstrate better NLL scores on both ID and OOD problem sets.}
\label{fig:NLL}     
\end{figure}

\begin{table*}[!t]
\small
\setlength\tabcolsep{4pt}
\begin{adjustbox}{width=\textwidth}
\begin{tabular}{l|l|cccc| cccc| cccc}
\hline
Noise  &\multirow{2}{*}{Method} &\multicolumn{4}{c}{AffectNet-C} &\multicolumn{4}{c}{AffWild-C} &\multicolumn{4}{c}{RAF-DB-C}\\
ratio& &Acc (Pre) &ECE (Post)  &Acc (Pre) &ECE (Post) &Acc (Pre) &ECE (Post)  &Acc (Pre) &ECE (Post) &Acc (Pre) &ECE (Post)  &Acc (Pre) &ECE (Post)\\
\hline \hline
&Baseline\cite{affectnet} &47.4 $\pm{0.9}$  &15.3 $\pm{2.5}$ &48.5 $\pm{0.1}$ &\underline{12.8 $\pm{1.5}$} &43.2 $\pm{0.3}$ &27.9 $\pm{5.0}$ &37.8 $\pm{2.9}$  &\underline{8.85 $\pm{0.3}$} &59.0 $\pm{0.7}$ &23.0 $\pm{0.9}$ &57.4 $\pm{0.9}$  &\underline{11.6 $\pm{0.6}$} \\
&SCN\cite{wang2020suppressing}  &46.8 $\pm{1.3}$  &13.3 $\pm{1.4}$ &47.1 $\pm{0.1}$ &\underline{12.5 $\pm{1.4}$} &43.2 $\pm{2.6}$ &26.4 $\pm{7.8}$ &41.9 $\pm{0.6}$  &\textbf{\underline{9.02 $\pm{0.9}$}} &60.4 $\pm{0.5}$ &24.3 $\pm{0.1}$ &60.0 $\pm{0.4}$  &\underline{12.3 $\pm{0.1}$} \\ 
10\% &DMUE\cite{she2021dive} &53.0 $\pm{0.5}$  &12.4 $\pm{0.5}$ &\textbf{\underline{53.6 $\pm{0.4}$}} &\textbf{\underline{9.39 $\pm{0.3}$}} &50.3 $\pm{0.7}$ &40.0 $\pm{1.8}$ &47.5 $\pm{1.3}$  &\underline{15.0 $\pm{3.4}$} &72.8 $\pm{0.3}$ &26.8 $\pm{1.7}$ &72.0 $\pm{0.5}$  &\underline{17.5 $\pm{3.5}$} \\
&RUL\cite{zhang2021relative} &51.7 $\pm{0.5}$  &12.7 $\pm{0.4}$ &51.4 $\pm{0.2}$ &\underline{12.4 $\pm{0.9}$} &50.1 $\pm{2.2}$ &18.9 $\pm{2.9}$ &\textbf{\underline{50.3 $\pm{0.2}$}}  &\underline{9.60 $\pm{1.7}$} &74.1 $\pm{0.1}$ &16.0 $\pm{0.3}$ &\textbf{\underline{74.6 $\pm{0.1}$}}  &\underline{12.5 $\pm{0.2}$} \\
&SWINV2\cite{liu2021swinv2} &51.6 $\pm{0.2}$  &\underline{11.0 $\pm{1.2}$} &50.9 $\pm{1.0}$ &11.6 $\pm{0.4}$ &48.6 $\pm{0.3}$ &23.2 $\pm{6.0}$ &44.7 $\pm{1.0}$  &\underline{7.03 $\pm{0.2}$} &70.9 $\pm{1.8}$ &16.2 $\pm{3.8}$ &72.5 $\pm{1.4}$  &\textbf{\underline{12.1 $\pm{1.8}$}} \\
\hline
 
&Baseline\cite{affectnet} &41.4 $\pm{2.9}$ &12.4 $\pm{2.5}$ &42.3 $\pm{2.6}$  &\underline{12.0 $\pm{3.2}$} &42.6 $\pm{2.5}$  &15.3 $\pm{1.3}$ &39.5 $\pm{1.9}$  &\underline{8.17 $\pm{0.7}$} &54.5 $\pm{0.8}$  &22.0 $\pm{1.1}$ &53.4 $\pm{1.4}$ &\textbf{\underline{11.4 $\pm{0.3}$}} \\
&SCN\cite{wang2020suppressing} &41.4 $\pm{1.7}$ &14.0 $\pm{0.6}$ &41.3 $\pm{0.6}$  &\textbf{\underline{11.0 $\pm{0.3}$}} &43.8 $\pm{3.8}$  &23.7 $\pm{9.6}$ &42.0 $\pm{3.5}$  &\underline{9.63 $\pm{0.3}$} &55.5 $\pm{0.6}$  &24.3 $\pm{0.5}$ &52.5 $\pm{0.3}$ &\underline{11.5 $\pm{0.8}$} \\
20\% &DMUE\cite{she2021dive} &46.1 $\pm{0.4}$ &15.4 $\pm{1.5}$ &\textbf{\underline{46.3 $\pm{1.0}$}}  &\underline{12.1 $\pm{1.0}$} &48.8 $\pm{1.5}$  &37.5 $\pm{1.0}$ &47.9 $\pm{1.6}$  &\underline{16.1 $\pm{5.2}$} &69.1 $\pm{0.7}$  &24.8 $\pm{4.5}$ &68.9 $\pm{1.5}$ &\underline{17.9 $\pm{4.4}$} \\
&RUL\cite{zhang2021relative} &46.2 $\pm{1.0}$ &13.7 $\pm{1.9}$ &45.5 $\pm{0.5}$  &\underline{13.6 $\pm{1.9}$} &\textbf{\underline{49.1 $\pm{0.8}$}}  &\underline{9.90 $\pm{0.2}$} &46.4 $\pm{0.2}$  &12.6 $\pm{0.3}$ &71.4 $\pm{0.2}$  &13.6 $\pm{0.1}$ &\textbf{\underline{72.2 $\pm{0.9}$}} &\underline{11.7 $\pm{0.2}$} \\
&SWINV2\cite{liu2021swinv2} &44.5 $\pm{0.9}$ &\underline{11.2 $\pm{0.4}$} &44.5 $\pm{0.2}$  &11.5 $\pm{1.3}$ &45.0 $\pm{2.0}$  &19.0 $\pm{3.9}$ &44.5 $\pm{0.3}$  &\underline{8.47 $\pm{0.6}$} &66.4 $\pm{0.3}$  &19.6 $\pm{6.4}$ &66.8 $\pm{1.3}$ &\underline{11.9 $\pm{2.8}$} \\
\hline

&Baseline\cite{affectnet} &36.3 $\pm{0.7}$ &15.2 $\pm{3.7}$ &35.8 $\pm{0.4}$ &\underline{13.0 $\pm{1.1}$} &37.2 $\pm{1.1}$ &\underline{8.10 $\pm{0.2}$} &37.5 $\pm{4.9}$ &10.1 $\pm{1.6}$ &51.0 $\pm{0.7}$ &24.5 $\pm{0.9}$ &52.3 $\pm{0.8}$ &\textbf{\underline{9.71 $\pm{0.7}$}} \\
&SCN\cite{wang2020suppressing} &37.4 $\pm{0.2}$ &14.6 $\pm{3.0}$ &36.1 $\pm{1.3}$ &\underline{10.8 $\pm{3.0}$} &38.9 $\pm{1.4}$ &17.3 $\pm{8.5}$ &37.5 $\pm{0.8}$ &\underline{11.8 $\pm{0.2}$} &53.7 $\pm{1.3}$ &25.2 $\pm{0.3}$ &49.0 $\pm{3.1}$ &\underline{10.3 $\pm{0.5}$} \\
30\% &DMUE\cite{she2021dive} &38.8 $\pm{0.6}$ &24.4 $\pm{9.9}$ &38.3 $\pm{0.4}$ &\underline{23.1 $\pm{3.6}$} &48.0 $\pm{4.7}$ &33.0 $\pm{6.5}$ &\textbf{\underline{48.1 $\pm{3.9}$}} &\underline{13.5 $\pm{3.2}$} &63.8 $\pm{4.2}$ &9.83 $\pm{1.4}$ &63.5 $\pm{4.3}$ &\underline{9.65 $\pm{0.1}$} \\
&RUL\cite{zhang2021relative} &38.2 $\pm{0.9}$ &14.8 $\pm{4.6}$ &36.7 $\pm{1.0}$ &\underline{13.8 $\pm{5.2}$} &47.6 $\pm{1.0}$ &\underline{8.95 $\pm{0.7}$} &45.5 $\pm{3.4}$ &13.5 $\pm{2.5}$ &66.9 $\pm{0.5}$ &13.6 $\pm{0.5}$ &\textbf{\underline{68.8 $\pm{0.1}$}} &\underline{11.5 $\pm{0.2}$} \\
&SWINV2\cite{liu2021swinv2} &\textbf{\underline{40.8 $\pm{0.5}$}} &\textbf{\underline{8.80 $\pm{0.1}$}} &40.6 $\pm{0.9}$ &9.52 $\pm{0.6}$ &44.2 $\pm{3.4}$ &12.4 $\pm{2.6}$ &44.6 $\pm{0.5}$ &\underline{11.4 $\pm{0.1}$} &61.0 $\pm{1.3}$ &19.5 $\pm{5.7}$ &63.6 $\pm{2.1}$ &\underline{10.3 $\pm{1.6}$} \\

\hline
\end{tabular}
\end{adjustbox}
\vspace{-3mm}
\caption{FER algorithms remains relatively accurate despite synthetic label noise but are uncalibrated. Even under the effects of label noise, our proposed strategy generally improves the calibration performance of all methods.}
\label{table:FER_C_cat_noise}
\end{table*}

\subsection{Entropy Analysis}
In \cref{fig:NLL} we illustrate the average NLL scores of all algorithms before and after calibration. As a statistical measure between a model's probabilities and the associated test labels, a lower NLL signifies a lower error between predictions and true values. For the ID evaluation, we measure the output probabilities of the uncalibrated and calibrated algorithms against the original one-hot labels. 

Apart from SWINV2 on AffectNet, all methods show high confidences and better NLL scores when trained with calibration, with an average improvement of (1.4\%, 31.4\%, 22.9\%) respectively for each dataset. For the OOD evaluation, we measure each model's probabilities against the proposed soft labels formulated in \cref{final_label}. Overall we see greater improvements of NLL scores (5.36\%, 33.5\%, 41.3\%) compared to the ID test sets. Our results suggests that calibrated FER models inherently adapt their confidences to ID and OOD settings, allowing models to remain confident ID and have higher entropy predictions OOD. 

\subsection{Experiments with Label Noise}
The key challenges to FER stems from visual ambiguity (i.e. subjective facial expressions and occlusions) and annotation disagreement (i.e. noisy labels). Since many FER algorithms are designed to handle mislabelled annotations, we further highlight the benefits of calibration by training with label noise ratios of (10\%, 20\% and 30\%). We follow the experiment settings of prior works \cite{wang2020suppressing, she2021dive, zhang2021relative, Le_2023_WACV, Neo_2023_CVPR} by randomly flipping a percentage of the training labels to other classes.

In Table \ref{table:FER_C_cat_noise} we show the mean and standard deviation from three repetitions of each method's OOD test accuracy and ECE scores (in \%) trained under label noise. We also compare the results when each method is trained with our proposed calibration strategy. Our findings show that although FER algorithms remain robust in terms of accuracy, they are not immune to miscalibration when trained under label noise. In most cases, calibration can be improved by applying our strategy, with better underlined ECE scores evaluated on \textsc{FER-C} despite the influences from label noise. We expected to see improvements in recognition accuracy with calibration considering how calibration can be viewed as a regularization technique. However, we do not find any major differences in accuracy - only improvements in ECE. Under increasing label noise, both RUL and DMUE consistently achieve the best performance consistently across datasets. For completeness, we also provide results of the ID label noise experiments in \cref{extra_results}.

\subsection{Calibration Component Analysis}
To investigate the effects of the proposed calibration strategy, we perform an ablation study of the following calibration components. We compare all five FER algorithms trained with different calibration loss function components on RAF-DB-C. Specifically, each method is trained with CE, Focal, MaxEnt and MaxEnt loss + MBLS. As shown in Table \ref{table:calibration_component_analysis}, CE loss results in the worst calibration performance, with an ECE of 29.8\%. By training each algorithm with Focal loss and MaxEnt loss, calibration performance is significantly improved by an overall average of 6.6\% and 8.72\%. The best results are achieved when models are trained using MaxEnt loss + MBLS, with an average improvement of about 1.08\% contributed from MBLS.

\begin{table}[!htb]
\begin{center}
\setlength\tabcolsep{2pt}
\begin{tabular}{lcccc}
\hline
Method & CE & Focal & MaxEnt & MaxEnt+MBLS \\
\hline \hline
Baseline  \cite{affectnet} &29.5 &20.0 &19.0 &\textbf{16.6} \\
SCN \cite{wang2020suppressing} &35.1 &22.0 &\textbf{19.8} &20.7   \\ 
DMUE \cite{she2021dive} &33.8 &29.0 &25.3 &\textbf{24.1} \\
%LDVLA \cite{Le_2023_WACV} &16.3 &19.5 &\textbf{12.2} &13.0\\
RUL \cite{zhang2021relative} &30.6 &\textbf{20.5} &24.8 &22.4\\
SwinV2-T \cite{liu2021swinv2} &33.6 &28.3 &25.4 &\textbf{23.2}\\
\hline \hline
Overall (Avg.) &29.8 &23.2 &21.08 &\textbf{20.0}\\
\hline
\end{tabular}
\end{center}
\vspace{-1.5em}
\caption{ECE (\%) for component analysis on RAF-DB-C. Uncalibrated models have the highest ECE, best results are achieved with MaxEnt loss + MBLS.}
\label{table:calibration_component_analysis}
\end{table}

\section{Conclusion}
%We presented a novel OOD FER benchmark complete with soft labels that reflect the uncertainty behind facial expressions and visual corruptions. We provide details of our soft image corruption framework along with their soft label formulation. Our benchmark features 85 corrupted forms of AffectNet, AffWild and RAF-DB. Our experiments illustrate the benefits of training well-calibrated FER models, with better ECE, KSE and NLL scores when evaluated ID and OOD. Calibration delivers meaningful probabilistic predictions and provides robustness against disturbances such as OOD shifted inputs and label noise. With no increased training times or sacrifice in accuracy, well-calibrated FER models are a vital step towards the safe, reliable deployment of FER models.
We presented \textsc{FER-C}, a novel OOD benchmark equipped with soft labels for ambiguous and corrupted FER. Our benchmark features 85 corrupted forms of AffectNet, AffWild and RAF-DB detailing their proposed soft image corruption framework and soft label formulation. Our experiments demonstrate the advantages of calibration, including meaningful probabilities and robustness against perturbations such as OOD-shifted inputs and label noise. Without incurring increased training times or compromising accuracy, well-calibrated models represent a crucial step towards the safe, reliable deployment of FER models.
%------------------------------------------------------------------------
%%%%%%%%% REFERENCES
%\clearpage
{
    \small
    \bibliographystyle{ieeenat_fullname}
    \bibliography{main}
}
\clearpage
\appendix
\section{Appendix}

\subsection{Additional Calibration Metrics}
\label{extra_metrics}
Apart from the calibration metrics described in the main paper, we further include other calibration metrics and discuss additional results here.

\noindent \textbf{Adaptive ECE (AdaECE)} is proposed to evenly measure samples across bins \cite{Nixon2019MeasuringCI} since the ECE is known to be biased towards bins with higher confidence intervals: AdaECE $= \sum^B_{b=1} \frac{n_b}{N} | \textrm{acc}(b) - \textrm{conf}(b) |$ s.t. $ \forall{b,i} \cdot |B_b| = |B_i| $.

\noindent \textbf{Classwise ECE (CECE):} The ECE only considers the $\max$ of the predicted confidences. For certain situations, it would be critical for the confidences of other classes to be well-calibrated. The CECE is then a useful metric that considers all K classes \cite{Nixon2019MeasuringCI}: 
\begin{equation}
CECE = \frac{1}{K} \sum^B_{b=1} \sum^{K}_{k=1} \frac{n_{b,k}}{N} | \textrm{acc}(b,k) - \textrm{conf}(b,k) |.
\end{equation}

\begin{figure*}[!htb]
    \includegraphics[width=\textwidth]{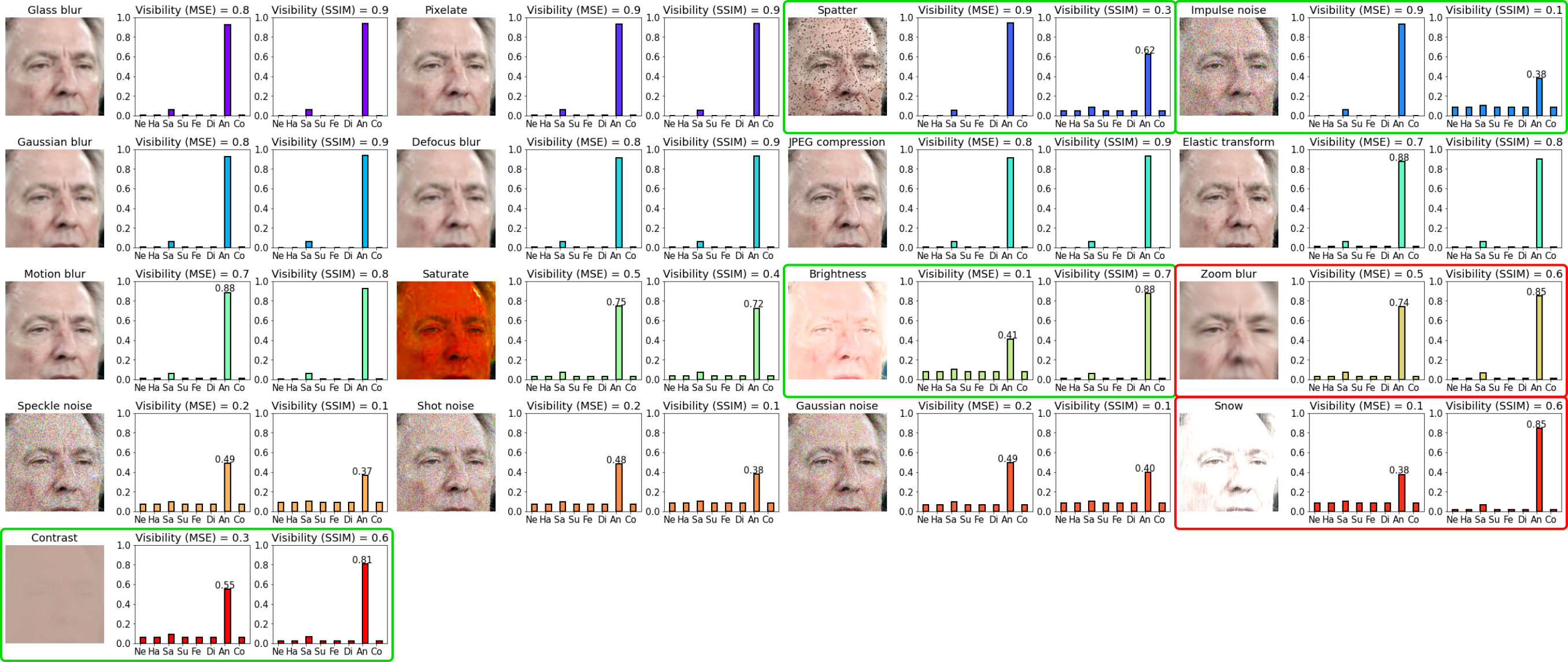}
    \vspace{-1em}
    \caption{In many cases, choosing between L2 error or SSIM for the visibility function gives similar results. In cases where results differ, SSIM performs better for certain corruptions (red), but L2 error paints a more accurate representation of the uncertainty for many other corruptions (lime). }
    \label{fig:soft_corruption_comparisons}
\end{figure*}

\subsection{Common Image Corruptions}
\label{corruption_details}
The 17 data augmentations proposed in the main paper aims to simulate the possible variations and intensities of OOD shifts in the real world. In essence, each of the 17 different corruptions can be summarized into four categories of augmentations namely: Noise, Blurs, Weathers and Digital. 

Specifically, Noise augmentations contain \textit{Gaussian}, \textit{Shot}, \textit{Impulse noise} and \textit{Speckle noise} which can be caused by electronic interference and bit errors. Blurs often come from camera movements and lens issues which includes \textit{Motion}, \textit{Defocus}, \textit{Glass}, \textit{Zoom} and \textit{Gaussian blur}. Influences from nature are included in Weather such as \textit{Snow}. Digital contains illumination changes and digital compression formats such as \textit{Brightness}, \textit{Contrast}, \textit{Elastic transform}, \textit{Pixelate}, \textit{JPEG compression}, \textit{Spatter} and \textit{Saturation}. 

We note that these augmentations merely represent a small subset of possible OOD shifts and that other image transformations and perturbations could be included in the future. However, we argue that these 85 corruptions are sufficiently diverse for evaluating FER models. In general, FER algorithms that provide improvements on this benchmark showcase robustness against OOD shifts and essential generalization properties necessary for deployment.

%In general, classifiers with lower overall calibration scores are considered to be better calibrated. We show the additional results using other calibration metrics for Synthethic OOD in \cref{table:augmented_extra} and Wilds in \cref{table:wilds_extra}. For the NLL and Brier score, we show the errors of misclassified samples as proposed by \cite{mukhoti2020calibrating}. Apart from metrics, calibration behaviour can also be visualized (see \cref{fig:probabilty_densities}) with bin-strength plots and reliability diagrams \cite{NiculescuMizil2005PredictingGP}. For reliability diagrams, perfectly calibrated models will have accuracy bins that equal their confidence.

\subsection{Image Quality Metrics}
\label{image_metrics}
In this subsection, we discuss alternative formulations of the visibility function for measuring image quality. Specifically, we compare between the L2 error and SSIM \cite{SSIM} between the clean source image $x_i$ and corrupted image $c(x_i)$. The SSIM was designed as a measure with regards to human perception, in contrast to L2 or L1 errors which measure the absolute pixelwise differences. The SSIM considers three quantities of human vision: luminance $L(x_i, c(x_i))$, contrast $C(x_i, c(x_i))$ and structure $S(x_i, c(x_i))$. The SSIM is then the product of the above three quantities with exponent constants ${\alpha, \beta, \gamma}$, and $\xi$ controlling the sliding window size:
\begin{equation}
    \text{SSIM}(x_i, c_i, \xi) = L(x, c)^\alpha \cdot C(x_i, c_i))^\beta \cdot S(x_i, c_i)^\gamma
\end{equation}

In \cref{fig:soft_corruption_comparisons}, we compare the qualitative differences between using L2 error and SSIM as our visibility function. In general, most corruptions yield similar soft labels when using either L2 error or SSIM \eg \textit{Glass blur}, \textit{Pixelate}, \textit{Gaussian blur}. For cases such as \textit{Snow}, SSIM give a better label representation due to its ability to capture the structure of the image. However, we find the L2 formulation to be more reliable in other cases where changes are either minor or extreme. For \eg in \textit{Contrast}, \textit{Spatter}, \textit{Impulse noise} we find that the labels for L2 error more accurate than that of using SSIM. Ultimately we find our formulation to be adequate approach for the proposed 17 corruptions. In cases where additional corruptions are to be introduced, a more thorough study and comparison of other image quality metrics may be required.

\begin{figure*}[!htb]
  \begin{subfigure}{0.33\textwidth}
    \includegraphics[width=\linewidth]{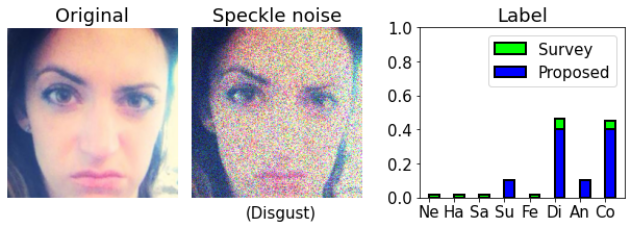}
    \caption{\textsc{AffectNet-C}} \label{fig:1a}
  \end{subfigure}%
  \begin{subfigure}{0.33\textwidth}
    \includegraphics[width=\linewidth]{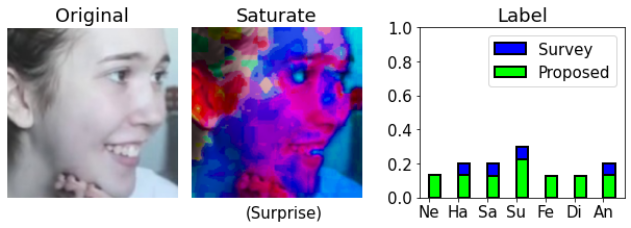}
    \caption{\textsc{AffWild-C}} \label{fig:1b}
  \end{subfigure}%
  \begin{subfigure}{0.33\textwidth}
    \includegraphics[width=\linewidth]{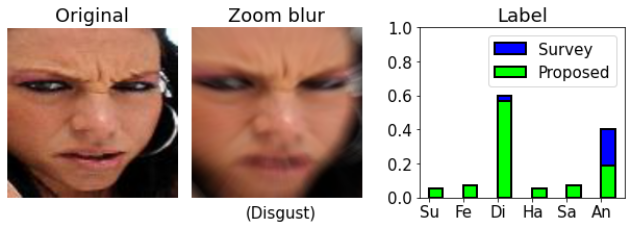}
    \caption{\textsc{RAF-DB-C}} \label{fig:1c}
  \end{subfigure}%
\vspace{-1.0em}
\caption{Comparisons between user surveyed labels (blue) and proposed labels (lime). Proposed labels better capture the underlying uncertainties and ambiguities of wild facial expressions than given one-hot labels.} 
\label{User_study}
\end{figure*}

\subsection{User Study}
\label{user_study}
To justify the proposed labels in our corrupted benchmark, we conduct a user study comprising of five FER experts. Each human expert is tasked to randomly survey 30 samples and each of their corrupted forms on all three datasets (Total 450 source images). In \cref{User_study} we show original samples (left) and a corrupted variant (centre) and the labels (right)  of \textsc{AffectNet-C}, \textsc{AffWild-C}, \textsc{RAF-DB-C}.

For the \textsc{AffWild-C} example shown in \cref{fig:1b}, we can see that the corruption \textit{Saturate} entirely distorts the original content of source image, making it extremely challenging to distinguish the original label \textit{Surprise}. Naturally, a highly corrupted image should have lower confidences as shown in our surveyed and proposed labels.

For both \cref{fig:1a} and \cref{fig:1c}, we can see that corruptions \textit{Speckle noise}, \textit{Zoom blur}, do not cause much information loss and retain high image visibility. For these cases, we expect test labels to have higher confidences, with our proposed labels better matching the surveyed labels. In general, our findings from our user-study show that our proposed labels are generally closer to the surveyed labels as compared to using one-hot labels.

\subsection{Implementation Details}
\label{implementation_details}
In our experiments, we follow the implementations provided by the respective authors \cite{wang2020suppressing, she2021dive, zhang2021relative, Neo_2023_CVPR} and use ResNet-18 \cite{he2015deep} as the backbone for our convolutional architectures. For vision transformers, we use SwinV2-T \cite{liu2021swinv2}. We referenced each author's code (if provided) and reproduced the experiments to the best of our ability. The parameter $\pi_k$ is computed from the class label frequency, $\mu_k$ and $\Sigma_k$ are computed from the VA values of the AffectNet training set.  

All input images are resized to $224 \times 224$ pixels with each algorithm trained using onehot labels and clean images (without image augmentations). For RAF-DB, we create onehot labels by taking the $\argmax$ of the given mixed labels. We follow the methodology of each author and optimize uncalibrated versions of each algorithm using cross-entropy (CE) loss and Adam \cite{kingma2017adam}. All backbone architectures are initialized with weights pre-trained on ImageNet, except DMUE which is initialized from pre-trained Celeb1M.

\subsection{Supplementary Results}
\label{extra_results}

\begin{table}[!htb]%\setlength\tabcolsep{0.25em}
\centering
\large
\begin{adjustbox}{width=\columnwidth}
\begin{tabular}{l|cc|cc|cc}
&\multicolumn{2}{c|}{(a) AffectNet} &\multicolumn{2}{c|}{(b) AffWild} &\multicolumn{2}{c}{(c) RAF-DB}\\
Method &CECE (Pre) &CECE (Post) &CECE (Pre) &CECE (Post) &CECE (Pre) &CECE (Post)\\
\hline
Baseline\cite{affectnet} &2.44 &3.36 &10.1 &8.14 &3.62 &8.54 \\
SCN\cite{wang2020suppressing} &2.30 &3.64 &10.4 &10.4 &4.61 &3.80 \\
DMUE\cite{she2021dive} &3.60 &2.10 &12.4 &20.1 &4.61 &3.76 \\
RUL\cite{zhang2021relative}  &2.57 &5.99 &8.15 &8.45 &2.97 &5.57 \\
SwinV2-T\cite{liu2021swinv2} &2.02 &11.2 &11.2 &10.4 &4.87 &4.11 \\
\hline
Overall (Avg.) &2.59 &5.25 &10.5 &11.5 &4.14 &5.16\ \\
&\multicolumn{2}{c|}{(d) AffectNet-C} &\multicolumn{2}{c|}{(e) AffWild-c} &\multicolumn{2}{c}{(f) RAF-DB-C}\\
Method &CECE (Pre) &CECE (Post) &CECE (Pre) &CECE (Post) &CECE (Pre) &CECE (Post)\\
\hline
Baseline\cite{affectnet} &20.5 &8.85 &25.7 &8.69 &21.5 &11.2 \\
SCN\cite{wang2020suppressing} &21.0 &10.8 &25.9 &10.6 &22.7 &14.7 \\
DMUE\cite{she2021dive} &22.5 &11.4 &28.1 &20.6 &22.00 &14.6 \\
RUL\cite{zhang2021relative}  &21.1 &10.8 &23.9 &7.81 &20.2 &12.8 \\
SwinV2-T\cite{liu2021swinv2} &19.7 &7.64 &26.7 &9.89 &21.6 &14.1 \\
\hline
Overall (Avg.) &20.9 &9.89 &26.1 &11.5 &21.6 &13.5 \\
\end{tabular}
\end{adjustbox}
\caption{Calibrated FER models are able to demonstrate better CECE scores on OOD problem sets.}
\label{table:cece}
\end{table}

\begin{table*}[!htb]%\setlength\tabcolsep{0.25em}
\centering
\large
\begin{adjustbox}{width=\textwidth}
\begin{tabular}{l|cccc|cccc|cccc}
&\multicolumn{4}{c|}{(a) AffectNet} &\multicolumn{4}{c|}{(b) AffWild} &\multicolumn{4}{c}{(c) RAF-DB}\\
Method &NLL (Pre) &AdaECE (Pre) &NLL (Post) &AdaECE (Post) &NLL (Pre) &AdaECE (Pre) &NLL (Post) &AdaECE (Post) &NLL (Pre) &AdaECE (Pre) &NLL (Post) &AdaECE (Post)\\
\hline
Baseline\cite{affectnet} &\underline{1.21 $\pm{0.1}$} &\underline{4.18 $\pm{0.9}$} &1.23 $\pm{0.1}$ &6.54 $\pm{2.6}$ &2.30 $\pm{0.2}$ &33.7 $\pm{1.7}$  &\underline{1.50 $\pm{0.1}$} &\underline{8.17 $\pm{0.2}$} &0.71 $\pm{0.1}$  &9.80 $\pm{0.4}$ &\underline{0.66 $\pm{0.1}$}  &\underline{7.50 $\pm{0.9}$} \\
SCN\cite{wang2020suppressing} &\underline{1.20 $\pm{0.1}$} &\underline{4.75 $\pm{2.9}$} &1.26 $\pm{0.1}$ &7.15 $\pm{3.2}$ &2.32 $\pm{0.1}$ &33.5 $\pm{0.9}$  &\underline{1.62 $\pm{0.1}$} &\underline{11.8 $\pm{0.9}$} &0.79 $\pm{0.1}$  &11.6 $\pm{1.6}$ &\underline{0.69 $\pm{0.1}$}  &\underline{5.01 $\pm{0.8}$} \\
DMUE\cite{she2021dive} &1.26 $\pm{0.1}$ &14.3 $\pm{2.4}$ &\underline{1.19 $\pm{0.1}$} &\underline{4.86 $\pm{2.4}$} &3.63 $\pm{0.1}$ &41.2 $\pm{0.8}$  &\underline{2.32 $\pm{0.1}$} &\underline{20.3 $\pm{0.5}$} &0.96 $\pm{0.1}$  &13.3 $\pm{0.1}$ &\underline{0.69 $\pm{0.1}$}  &\underline{6.27 $\pm{1.8}$} \\
RUL\cite{zhang2021relative} &1.23 $\pm{0.1}$ &7.03 $\pm{0.1}$ &\underline{1.23 $\pm{0.1}$} &\underline{5.79 $\pm{0.2}$} &1.58 $\pm{0.1}$ &22.6 $\pm{0.4}$  &\underline{1.31 $\pm{0.1}$} &\underline{9.60 $\pm{0.8}$} &0.65 $\pm{0.1}$  &5.98 $\pm{1.7}$ &\underline{0.61 $\pm{0.1}$}  &\underline{4.86 $\pm{0.9}$} \\
SwinV2-T\cite{liu2021swinv2} &\underline{1.15 $\pm{0.1}$} &\underline{3.61 $\pm{0.7}$} &1.19 $\pm{0.1}$ &9.76 $\pm{1.4}$ &2.63 $\pm{0.1}$ &39.0 $\pm{1.1}$  &\underline{1.67 $\pm{0.1}$} &\underline{13.2 $\pm{1.2}$} &0.94 $\pm{0.1}$  &13.8 $\pm{0.2}$ &\underline{0.66 $\pm{0.1}$}  &\underline{4.91 $\pm{0.4}$} \\
&\multicolumn{4}{c}{(d) AffectNet-C} &\multicolumn{4}{c}{(e) AffWild-C} &\multicolumn{4}{c}{(f) RAF-DB-C}\\
Method &NLL (Pre) &AdaECE (Pre) &NLL (Post) &AdaECE (Post) &NLL (Pre) &AdaECE (Pre) &NLL (Post) &AdaECE (Post) &NLL (Pre) &AdaECE (Pre) &NLL (Post) &AdaECE (Post)\\
\hline
Baseline\cite{affectnet} &1.99 $\pm{0.1}$ &12.3 $\pm{0.9}$  &\underline{1.87 $\pm{0.1}$} &\underline{8.50 $\pm{0.5}$} &3.23 $\pm{0.2}$ &32.69 $\pm{0.4}$  &\underline{2.10 $\pm{0.1}$} &\underline{10.1 $\pm{1.1}$} &2.22 $\pm{0.1}$ &23.1 $\pm{0.4}$  &1\underline{.49 $\pm{0.1}$} &\underline{11.2 $\pm{0.1}$} \\
SCN\cite{wang2020suppressing} &1.99 $\pm{0.1}$ &12.6 $\pm{2.8}$  &\underline{1.92 $\pm{0.1}$} &\underline{9.39 $\pm{1.5}$} &3.38 $\pm{0.1}$ &32.81 $\pm{0.6}$  &\underline{2.21 $\pm{0.1}$} &\underline{12.5 $\pm{1.6}$} &2.44 $\pm{0.3}$ &24.9 $\pm{2.4}$  &\underline{1.58 $\pm{0.1}$} &\underline{13.3 $\pm{1.4}$} \\
DMUE\cite{she2021dive} &2.55 $\pm{0.1}$ &22.0 $\pm{1.6}$  &\underline{2.24 $\pm{0.1}$} &\underline{13.3 $\pm{1.9}$} &5.51 $\pm{0.1}$ &41.92 $\pm{0.3}$  &\underline{3.73 $\pm{0.1}$} &\underline{20.5 $\pm{0.4}$} &3.09 $\pm{0.1}$ &25.4 $\pm{0.1}$  &\underline{1.93 $\pm{0.3}$} &\underline{17.5 $\pm{2.6}$} \\
RUL\cite{zhang2021relative} &1.88 $\pm{0.1}$ &15.7 $\pm{0.1}$  &\underline{1.76 $\pm{0.1}$} &\underline{10.7 $\pm{0.2}$} &2.35 $\pm{0.1}$ &26.28 $\pm{0.7}$  &\underline{1.90 $\pm{0.1}$} &\underline{8.25 $\pm{0.2}$} &1.65 $\pm{0.2}$ &18.1 $\pm{2.1}$  &\underline{1.30 $\pm{0.1}$} &\underline{13.2 $\pm{0.4}$} \\
SwinV2-T\cite{liu2021swinv2} &1.88 $\pm{0.1}$ &11.3 $\pm{1.7}$  &\underline{1.78 $\pm{0.1}$} &\underline{7.72 $\pm{0.1}$} &3.98 $\pm{0.1}$ &37.58 $\pm{0.6}$  &\underline{2.37 $\pm{0.1}$} &\underline{12.4 $\pm{1.5}$} &2.69 $\pm{0.1}$ &24.2 $\pm{0.1}$  &\underline{1.45 $\pm{0.1}$} &\underline{14.7 $\pm{0.6}$} \\
\hline
\end{tabular}
\end{adjustbox}
\vspace{-3mm}
\caption{
Additional evaluation results (\%) for ID (top) and OOD (bottom) test sets over 3 random seeds. Calibration largely benefits NLL and AdaECE on both ID and OOD problem sets.}
\label{table:supp}
\end{table*}

\begin{table*}[!tb]
\small
\setlength\tabcolsep{4pt}
\begin{adjustbox}{width=\textwidth}
\begin{tabular}{l|l|cccc| cccc| cccc}
\hline
Noise  &\multirow{2}{*}{Method} &\multicolumn{4}{c}{AffectNet} &\multicolumn{4}{c}{AffWild} &\multicolumn{4}{c}{RAF-DB}\\
ratio& &Acc (Pre) &ECE (Post)  &Acc (Pre) &ECE (Post) &Acc (Pre) &ECE (Post)  &Acc (Pre) &ECE (Post) &Acc (Pre) &ECE (Post)  &Acc (Pre) &ECE (Post)\\
\hline \hline
&Baseline\cite{affectnet} &53.8 $\pm{0.9}$ &\underline{13.7 $\pm{4.7}$} &54.2 $\pm{0.7}$ &18.8 $\pm{0.2}$ &48.5 $\pm{0.5}$ &21.2 $\pm{9.9}$ &47.1 $\pm{1.9}$ &\underline{5.71 $\pm{0.9}$} &74.0 $\pm{0.2}$ &9.95 $\pm{2.0}$ &73.6 $\pm{1.1}$ &\underline{9.09 $\pm{0.3}$} \\
&SCN\cite{wang2020suppressing} &52.8 $\pm{1.1}$ &\underline{14.2 $\pm{1.7}$} &53.3 $\pm{0.4}$ &16.8 $\pm{0.8}$ &50.4 $\pm{1.9}$ &25.0 $\pm{9.9}$ &44.1 $\pm{3.0}$ &\underline{7.81 $\pm{0.4}$} &76.5 $\pm{0.1}$ &11.8 $\pm{1.7}$ &73.3 $\pm{0.7}$ &\underline{3.99 $\pm{0.4}$} \\
10\% &DMUE\cite{she2021dive} &56.9 $\pm{0.7}$ &6.10 $\pm{0.1}$ &57.8 $\pm{0.3}$ &\underline{5.92 $\pm{0.2}$} &50.0 $\pm{1.6}$ &38.7 $\pm{0.8}$ &47.6 $\pm{0.6}$ &\underline{14.9 $\pm{1.1}$} &77.3 $\pm{0.4}$ &15.8 $\pm{2.5}$ &76.1 $\pm{0.3}$ &\underline{5.80 $\pm{3.8}$} \\
&RUL\cite{zhang2021relative} &55.4 $\pm{0.5}$ &\underline{11.6 $\pm{0.4}$} &55.0 $\pm{0.1}$ &11.9 $\pm{0.7}$ &51.2 $\pm{0.5}$ &\underline{7.39 $\pm{2.6}$} &50.8 $\pm{2.0}$ &10.6 $\pm{0.1}$ &77.6 $\pm{0.1}$ &\underline{4.70 $\pm{0.1}$} &78.4 $\pm{0.1}$ &8.30 $\pm{0.1}$ \\
&SWINV2\cite{liu2021swinv2} &54.9 $\pm{0.1}$ &\underline{11.6 $\pm{0.1}$} &54.5 $\pm{1.1}$ &14.9 $\pm{0.3}$ &48.4 $\pm{2.3}$ &22.6 $\pm{3.8}$ &46.3 $\pm{1.7}$ &\underline{5.95 $\pm{0.1}$} &75.6 $\pm{0.3}$ &7.20 $\pm{3.1}$ &77.0 $\pm{0.6}$ &\underline{5.20 $\pm{1.4}$} \\
\hline
 
&Baseline\cite{affectnet} &47.3 $\pm{3.6}$ &11.9 $\pm{5.1}$ &48.0 $\pm{3.6}$ &\underline{11.7 $\pm{7.0}$} &48.7 $\pm{1.5}$ &\underline{3.76 $\pm{1.6}$} &44.2 $\pm{1.3}$ &9.51 $\pm{1.8}$ &70.1 $\pm{0.5}$ &9.78 $\pm{0.1}$ &69.9 $\pm{1.0}$ &\underline{7.56 $\pm{1.9}$} \\
&SCN\cite{wang2020suppressing} &45.0 $\pm{2.1}$ &\underline{9.48 $\pm{1.0}$} &47.9 $\pm{0.1}$ &16.1 $\pm{1.0}$ &45.4 $\pm{1.1}$ &22.4 $\pm{9.9}$ &45.9 $\pm{2.0}$ &\underline{9.44 $\pm{1.6}$} &71.5 $\pm{0.1}$ &12.2 $\pm{0.3}$ &69.5 $\pm{0.3}$ &\underline{6.68 $\pm{0.9}$} \\
20\% &DMUE\cite{she2021dive} &49.6 $\pm{0.7}$ &8.10 $\pm{2.4}$ &49.7 $\pm{1.5}$ &\underline{4.61 $\pm{1.5}$} &45.8 $\pm{1.9}$ &37.1 $\pm{0.3}$ &47.9 $\pm{0.8}$ &\underline{14.3 $\pm{5.4}$} &73.2 $\pm{0.2}$ &14.0 $\pm{5.9}$ &71.9 $\pm{1.0}$ &\underline{7.32 $\pm{5.8}$} \\
&RUL\cite{zhang2021relative} &50.3 $\pm{0.9}$ &\underline{13.8 $\pm{2.9}$} &49.4 $\pm{0.4}$ &14.7 $\pm{2.1}$ &50.6 $\pm{0.4}$ &\underline{4.92 $\pm{0.1}$} &50.0 $\pm{0.7}$ &16.5 $\pm{1.6}$ &75.1 $\pm{0.3}$ &\underline{4.00 $\pm{0.2}$} &76.1 $\pm{1.0}$ &9.17 $\pm{0.7}$ \\
&SWINV2\cite{liu2021swinv2} &47.8 $\pm{0.4}$ &\underline{10.8 $\pm{1.5}$} &47.8 $\pm{0.3}$ &13.2 $\pm{2.1}$ &46.7 $\pm{1.0}$ &12.3 $\pm{2.9}$ &47.7 $\pm{1.7}$ &\underline{9.70 $\pm{0.9}$} &70.9 $\pm{0.5}$ &\underline{13.6 $\pm{4.1}$} &71.8 $\pm{0.7}$ &14.7 $\pm{7.4}$ \\
\hline

&Baseline\cite{affectnet} &42.5 $\pm{1.2}$ &12.0 $\pm{1.8}$ &41.7 $\pm{1.1}$ &\underline{9.43 $\pm{2.1}$} &43.6 $\pm{1.9}$  &\underline{5.45 $\pm{1.1}$} &41.27 $\pm{8.1}$ &13.7 $\pm{5.7}$ &66.3 $\pm{0.5}$ &12.8 $\pm{1.5}$ &67.5 $\pm{0.8}$ &\underline{11.1 $\pm{0.2}$} \\
&SCN\cite{wang2020suppressing} &43.5 $\pm{0.3}$ &\underline{9.25 $\pm{1.2}$} &41.8 $\pm{2.4}$ &12.4 $\pm{0.3}$ &43.8 $\pm{0.8}$  &\underline{12.7 $\pm{9.9}$} &46.59 $\pm{2.9}$ &13.7 $\pm{5.3}$ &68.7 $\pm{0.5}$ &13.6 $\pm{0.1}$ &63.8 $\pm{1.1}$ &\underline{5.99 $\pm{0.1}$} \\
30\% &DMUE\cite{she2021dive} &41.7 $\pm{1.0}$ &21.5 $\pm{9.9}$ &41.3 $\pm{0.2}$ &\underline{19.8 $\pm{5.2}$} &50.7 $\pm{3.9}$  &20.9 $\pm{2.4}$ &54.58 $\pm{0.6}$ &\underline{7.58 $\pm{3.5}$} &66.9 $\pm{3.4}$ &\underline{6.08 $\pm{3.1}$} &66.8 $\pm{3.7}$ &10.7 $\pm{2.6}$ \\
&RUL\cite{zhang2021relative} &41.3 $\pm{1.0}$ &11.8 $\pm{3.5}$ &40.7 $\pm{1.0}$ &\underline{9.08 $\pm{4.2}$} &48.5 $\pm{0.4}$  &\underline{5.46 $\pm{0.3}$} &45.93 $\pm{2.7}$ &13.2 $\pm{3.1}$ &70.0 $\pm{0.2}$ &\underline{5.09 $\pm{0.2}$} &72.3 $\pm{0.5}$ &9.79 $\pm{0.2}$ \\
&SWINV2\cite{liu2021swinv2} &44.2 $\pm{0.9}$ &\underline{7.92 $\pm{0.7}$} &43.9 $\pm{1.2}$ &10.9 $\pm{0.7}$ &47.0 $\pm{2.5}$  &\underline{4.00 $\pm{2.1}$} &45.83 $\pm{1.1}$ &11.9 $\pm{1.2}$ &65.0 $\pm{2.1}$ &18.2 $\pm{0.1}$ &68.3 $\pm{1.4}$ &\underline{4.78 $\pm{1.5}$} \\
\hline

\end{tabular}
\end{adjustbox}
\vspace{-3mm}
\caption{ID experiment results of FER algorithms trained under label noise. While calibration can enhance the performance of FER algorithms in some instances, it has a more pronounced impact on improving OOD performance}
\label{table:ID_FER_C_cat_noise}
\end{table*}

Apart from the main results discussed in the main paper, we further report additional results of each algorithm with and without calibration evaluated using AdaECE and the raw NLL scores reported in \cref{fig:NLL} of the main paper. We also report the CECE (\%) scores and the in-distribution results of each algorithm trained with and without calibration under varying levels of label noise evaluated on the original test sets over three random seeds. 

\subsubsection{Additional Calibration Results}
Regarding classwise calibration, we observe in Table \ref{table:cece} that calibration tends to greatly improve the CECE scores of each algorithm OOD. For ID results, our findings suggests that our proposed calibration strategy requires a trade-off between ID and OOD scores. Indeed, we ideally wish for all models to perform well regardless of ID or OOD, however this task is non-trivial and requires further work.

In Table \ref{table:supp}, our findings generally follow the consensus reported in the main results of Table \ref{table:main}. For ID experiments, we observe that calibration is able to improve NLL and AdaECE scores of each algorithm with the exception of the Baseline, SCN and SWINV2 on AffectNet. For these algorithms, our calibration strategy does not improve the ID performance, with slight decrements in calibration performance. For OOD results, we observe clear benefits from calibration with consistent improvements in NLL and AdaECE scores for all algorithms across all three datasets.

\subsubsection{ID Experiments with Label Noise}
In the main paper, we discussed the performance of each algorithm trained under varying level of synthetic label noise and reported the results evaluated on FER-C. Here we further report the ID results of each algorithm evaluated on the original test set of each dataset (clean images and onehot labels). 

In most cases, we find that calibration still benefits ID testing, with many algorithm performing well after calibration on datasets such as RAf-DB and AffWild. For AffectNet we observe a mixed performance in results, with less consistent behaviour than evaluated on the OOD set. Currently, we suspect that the main reason for the decrease in performance is due to the chosen calibration strategy in \cref{final_objective_func}, which is specifically designed for OOD problem sets. As discussed in the main paper, choosing an optimal calibration objective function that works well on both ID and OOD problems is non-trivial and requires future work.

\subsection{Discussion and Future Work}
\textbf{Progressive benchmark:} While we focused on FER image classification tasks, we believe that it is possible to extend out framework to include other FER datasets such as those collected in laboratories or video datasets. Other image corruptions and perturbations such as rotations, translations and affinities can also be incorporated into future designs.

\noindent \textbf{Label Formulation:} As we discussed the limitations of our soft label formulation in \cref{fig:soft_corruption_comparisons}. We would like to highlight that it is possible to design other interpretations of \cref{visibility} such as those using SSIM, PSNR etc. The visibility function can also include a combination or careful selection of different image quality measurements, such as using different visibilty functions based on different corruptions. We also believe that our proposed label formulation can be used during training as per \cite{Liu_2023_CVPR} to improve FER model recognition performance.

\noindent \textbf{Calibrated FER:} We discussed the different performance of FER with and without calibration on ID and OOD problems. For future work, we believe that the study of different effects of calibration on ID and OOD problems set an interesting direction for further study. Ultimately, we believe that our benchmark opens up more future works for both FER and calibration fronts - which is an important step towards the deployment of safe, robust and reliable FER models.

\end{document}